\documentclass[10pt,twocolumn,letterpaper]{article}

\usepackage{import}
\usepackage[accsupp]{axessibility}
\usepackage{microtype}
\usepackage{graphicx}

\usepackage{xspace} 

\usepackage{amsmath,amssymb,amsfonts}
\usepackage{etoolbox}

\usepackage[thmmarks, amsmath, thref]{ntheorem}
\usepackage{bbm}
\usepackage{bm}

\usepackage[normalem]{ulem} 

\usepackage{times}
\usepackage{epsfig}
\usepackage{graphicx}
\usepackage{subcaption}

\usepackage{diagbox}

\usepackage{lipsum}  

\usepackage{threeparttable}
\usepackage[singlelinecheck=true,justification=centering]{caption}
\usepackage{makecell}
\usepackage{multirow}
\usepackage{color,soul}
\usepackage{xcolor}  

\definecolor{dimgray}{rgb}{0.35, 0.35, 0.35}
\usepackage{pifont}

\subimport{./}{macros}

\usepackage{cvpr}              
\usepackage{graphicx}
\usepackage{amsmath}
\usepackage{amssymb}
\usepackage{booktabs}

\usepackage[pagebackref=true,breaklinks=true,letterpaper=true,colorlinks,bookmarks=false,linkcolor=blue,urlcolor=blue]{hyperref}
\newcommand{\footnotestandalone}[1]{\let\thefootnote\relax\footnotetext{#1}}

\usepackage[capitalize]{cleveref}

\crefname{section}{Sec.}{Secs.}
\Crefname{section}{Section}{Sections}
\Crefname{table}{Table}{Tables}
\crefname{table}{Tab.}{Tabs.}

\def\cvprPaperID{4337}
\def\confName{CVPR}
\def\confYear{2022}

\begin{document}

\title{DeepDPM: Deep Clustering With an Unknown Number of Clusters}

\author{Meitar Ronen \hspace{10em} Shahaf E. Finder \hspace{10em} Oren Freifeld\\
The Department of Computer Science, Ben-Gurion University of the Negev\\
{\tt\small meitarr@post.bgu.ac.il \hspace{8em} \tt\small finders@post.bgu.ac.il \hspace{8em} \tt\small orenfr@cs.bgu.ac.il}
}

\maketitle

\begin{abstract}
Deep Learning (DL) has shown great promise in the unsupervised task of clustering.  That said, while in classical (\ie, non-deep) clustering the benefits of the nonparametric approach are well known, most deep-clustering methods are parametric: namely, they require a predefined and fixed number of clusters, denoted by $K$. When $K$ is unknown, however, using model-selection criteria to choose its optimal value might become computationally expensive, especially in DL as the training process would have to be repeated numerous times. In this work, we bridge this gap by introducing an effective deep-clustering method that does not require knowing the value of $K$ as it infers it during the learning. Using a split/merge framework, a dynamic architecture that adapts to the changing $K$, and a novel loss, our proposed method outperforms existing  nonparametric methods (both classical and deep ones). While the very few existing deep nonparametric methods lack scalability, we demonstrate ours by being the first to report the performance of such a method on ImageNet. We also demonstrate the importance of inferring $K$ by showing how methods that fix it deteriorate in performance when their assumed $K$ value gets further from the ground-truth one, especially on imbalanced datasets. Our code is available at~\url{https://github.com/BGU-CS-VIL/DeepDPM}.

\end{abstract}

\footnotestandalone{\small {\textbf{Acknowledgements.}
This work was supported by the Lynn and William Frankel Center at BGU CS, by the Israeli Council for Higher Education via the BGU Data Science Research Center, and by Israel Science Foundation Personal Grant \#360/21. M.R.~was also funded 
by the VATAT National excellence scholarship for female
Master’s students in Hi-Tech-related fields.}
}
\section{Introduction}\label{Sec:Intro}
\begin{figure}[t!]
\centering
\vspace{-.6cm}
	\subcaptionbox{ImageNet50: The original balanced dataset \label{balanced}}
		{\includegraphics[scale=.36]{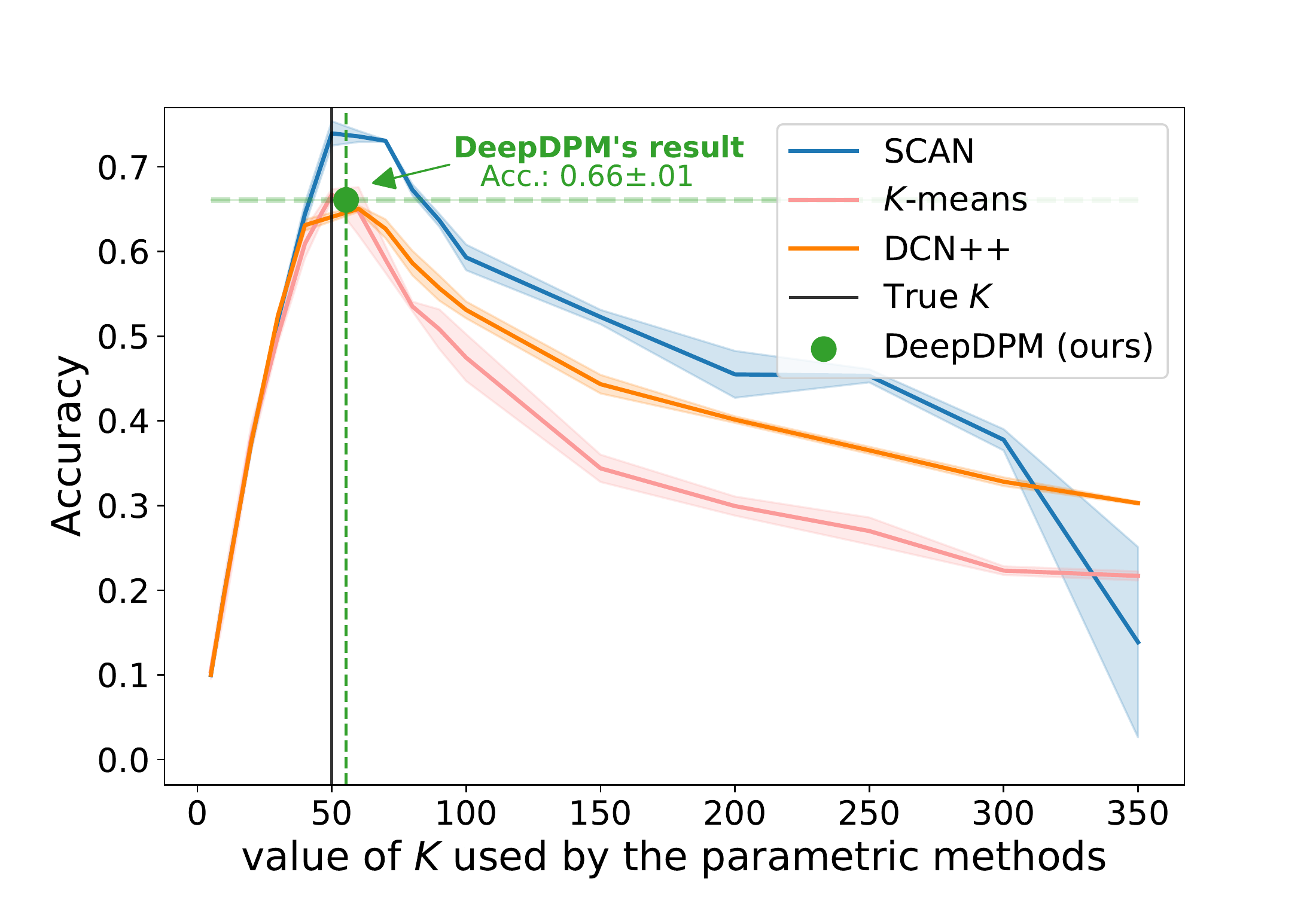}\vspace{-.05cm}}
\WHITE{--------} \vspace{-.75cm} \newline
	\subcaptionbox{ImageNet50: An imbalanced dataset \label{imbalanced}}
		{\includegraphics[scale=.36]{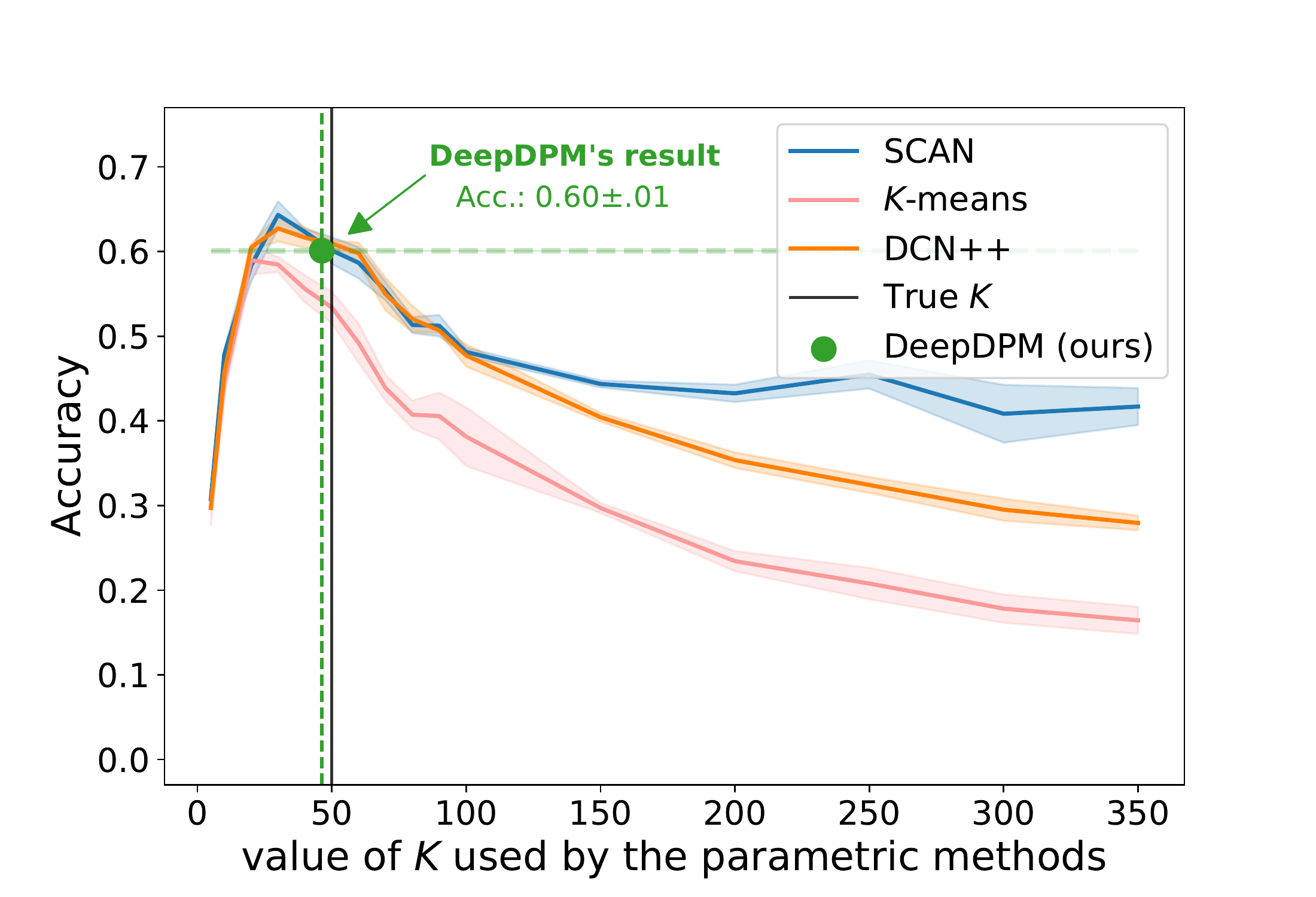}\vspace{-.05cm}}
\WHITE{--------} \vspace{-.75cm} \newline
\captionsetup{justification=justified, singlelinecheck=false}
\caption{Mean clustering accuracy of 3 runs ($\pm$ std.~dev.) on ImageNet50.
The Ground Truth $K$ is 50. 
Parametric methods such as $K$-means, DCN++ (an improved variant of~\cite{Yang:ICML:2017:towards})
and SCAN~\cite{Gansbeke:ECCV:2020:SCAN}, require knowing $K$. When given a poor estimate of $K$, they deteriorate in performance in a balanced dataset (a) and even more so in an imbalanced dataset (b). In contrast, the proposed DeepDPM does not require knowing $K$ (it infers its value; \eg, $K=55.3\pm1.53$ in (a) and $46.3\pm2.52$ in (b)) and yet yields comparable results. 
}
\label{results_figs:balanced_diff_K}
\end{figure}

Clustering is an important unsupervised-learning task where, unlike in the supervised case of classification, class labels are unavailable. Moreover, in the purely-unsupervised (and more realistic) setting this work focuses on, the number of classes, denoted by $K$, and their relative sizes (\ie, the class weights) are unknown too. 

The emergence of Deep Learning (DL) has not skipped clustering tasks. 
DL methods usually cluster large and high-dimensional datasets better and more efficiently than classical (\ie, non-deep) clustering methods~\cite{Yang:ICML:2017:towards,Gansbeke:ECCV:2020:SCAN}. That said, while in  classical clustering it is well understood that \emph{nonparametric} methods (namely, methods that find $K$) 
have advantages over \emph{parametric} ones (namely, methods that require a known $K$)~\cite{Sudderth:PhD:2006:GraphicalModels,Chang:Thesis:2014:Sampling}, there are only a few nonparametric deep clustering methods. Unfortunately, the latter are neither scalable nor 
effective enough. \textit{Our work bridges this gap by proposing an effective deep nonparametric method}, called DeepDPM.
In fact, even when $K$ is known, DeepDPM still achieves
results comparable to leading parametric methods (especially in imbalanced cases) despite their ``unfair'' advantage; see, \eg,~\autoref{results_figs:balanced_diff_K} or~\autoref{Sec:Results}.

More generally, the ability to infer the latent $K$ has \emph{practical} benefits,
including the following ones. 
1) Without a good estimate of $K$, parametric methods might suffer in performance. \autoref{results_figs:balanced_diff_K} shows that using the wrong $K$ 
can have a significant negative effect on parametric methods in both balanced and imbalanced datasets. When the value of $K$ becomes more and more inaccurate, even a State-Of-The-Art (SOTA) parametric deep clustering method, SCAN~\cite{Gansbeke:ECCV:2020:SCAN}, deteriorates in performance significantly. 
2) Changing $K$ during training has positive optimization-related implications; \eg, by splitting a single cluster into two, multiple data labels are changed \emph{simultaneously}. This often translates to large moves on the optimization surface which may lead to convergence to better local optima and performance gains~\cite{Chang:NIPS:2013:ParallelSamplerDP}; \eg, in~\autoref{Sec:Results} we demonstrate cases where nonparametric methods, ours included, outperform parametric ones even when the latter are given the true $K$.
3) A common workaround to not knowing $K$ is to use \emph{model selection}: namely, run a parametric method numerous times, using different $K$ values over a wide range, and then choose the ``best'' $K$ via an unsupervised criterion. 
That approach, however, besides missing the aforementioned potential gains (not being able to make large moves), does not scale and is usually infeasible for large datasets, \emph{especially in DL}. Moreover, \emph{the negative societal impact of the model-selection approach must be noted as well}: training a deep net tens or hundreds of times on a large dataset consumes prohibitively-large amounts of energy. 
4) $K$ itself may be a sought-after quantity of importance.

Bayesian nonparametric (BNP) mixture models, exemplified by the Dirichlet Process Mixture (DPM) model, offer an elegant, data-adaptive, and mathematically-principled solution for clustering when $K$ is unknown. However, the high computational cost typically associated with DPM inference is arguably why only a few works tried to use it in conjunction with deep clustering (\eg,~\cite{DC_BNP:Chen:2015:deepbeliefBNP, DC_BNP:Wang:2021:BNP_dnb, DC_BNP:Zhao:2019:adapVAE}).
Here we propose to \emph{combine the benefits of DL and the DPM effectively}. The proposed method, DeepDPM, uses splits and merges of clusters to change $K$ together with a dynamic architecture to accommodate for such changes. It also uses a novel amortized inference for Expectation-Maximization (EM) algorithms in mixture models. DeepDPM can be incorporated in deep pipelines that rely on clustering (\eg, for feature learning).
Unlike an offline clustering step (\eg, $K$-means), DeepDPM is differentiable during most of the training (the exception is during the discrete splits/merges) and thus supports gradient propagation through it. 
DeepDPM outperforms existing 
nonparametric clustering methods (both classical and deep ones) across several datasets and metrics. It also handles class imbalance gracefully and scales well to large datasets.
While we focus on clustering and not feature learning, we also show 
examples of clustering on pretrained features
as well as jointly learning features and clustering in an end-to-end fashion. 
To summarize, \textbf{our key contributions are:}
1) A deep clustering method that infers the number of clusters.
2) A novel loss that enables a new amortized inference in mixture models.
3) A demonstration of the importance, in deep clustering, of inferring $K$.   
4) Our method outperforms existing nonparametric clustering methods 
and we are the first
to report results of a deep nonparametric clustering method on 
a large dataset such as ImageNet~\cite{Deng:CVPR:2009:imagenet}.
\section{Related Work}\label{Sec:Related}

\textbf{Parametric Deep Clustering methods.} Recent such works can be divided into two types: two-step approaches and end-to-end ones. In the former, clustering is performed on features extracted in a pretext task.
For instance, Mc-Conville \etal~\cite{DC:Mcconville:ICPR:2021:n2d} run $K$-means on the embeddings, transformed by UMAP~\cite{DC:Mcinnes:2018:umap}, of a pretrained Autoencoder (AE).
While not scalable,~\cite{DC:Mcconville:ICPR:2021:n2d} achieves competitive results when it is applicable. Another  example is SCAN~\cite{Gansbeke:ECCV:2020:SCAN} which uses unsupervised pretrained feature extractors (\eg, MoCo~\cite{DC:Chen:2020:MoCO} and SimCLR~\cite{DC:Chen:ICML:2020L:simclr}). While reaching SOTA results, SCAN, being parametric, depends on having an estimate of $K$ and, as we show,  deteriorates in performance when the estimate is too inaccurate. Moreover, SCAN assumes uniform class weights (\ie a balanced dataset)
and that is often unrealistic in purely-unsupervised cases. 

End-to-end deep methods jointly learn features and clustering, possibly by alternation. Several works use an AE, or a Variational AE (VAE), with an additional clustering loss~\cite{Yang:ICML:2017:towards, DC:Xie:ICLR:2016:xie_DC, DC:Wu:2020:deep_ae_geo, DC:Yang:ICCV:2019:deepGMM, DC:Jiang:2016:VaDE}; \eg, DCN~\cite{Yang:ICML:2017:towards} runs $K$-means on the embeddings of a pretrained AE, and retrains it with a loss consisting of a reconstruction term and a clustering-based term, to simultaneously update the features, clusters' centers, and assignments.
Other works, \eg~\cite{Caron:2019:ICCV:unsupervised,Caron:2018:ECCV:deepcluster}, 
use 
convolutional neural nets to alternately learn features and clustering.

While our work focuses on clustering, not feature learning, we demonstrate how it can also be incorporated with the two approaches above. 
Moreover, all the methods above assume a predefined and fixed $K$ and, at least the more effective ones among them, take substantial time and resources
to train (so searching for the ``right'' $K$ using model selection is costly and/or inapplicable).

\textbf{Nonparametric Classical Clustering.}
Closely related to our work is BNP clustering and, more specifically, the DPM model~\cite{Ferguson:1973:BNP,Antoniak:1974:DPMM}. While many computer-vision works  rely on BNP clustering
\cite{Sudderth:PhD:2006:GraphicalModels,Jian:ICCV:2007:DPM,Kivinen:ICCV:2007:Multiscale,Sudderth:NIPS:2008:Shared,Sudderth:IJCV:2008d:Describing,Gomes:2008:CVPR:incremental,Li:IJCV:2010:Optimol,Cherian:CVPR:2011:DPSPD,Hughes:CVPRw:2012:BNP,Ghosh:CVPR:2012:nonparametric,Ghosh:NIPS:2012:BNPdef2parts,Hu:PAMI:2013:incrementalDPMMTrajectoryClustering,Chang:ECCV:2014:Intrinsic,Ghosh:UAI:2014:ddHier,neiswanger:AIS:2014:dependentDPMMDetectionFreeTracking,Straub:CVPR:2015:SmallVar,Steinberg:CVIU:2015:HierBaysianScene,Cabezas:ICCV:2015:Semantic,Ji:ICML:2017:BNPpatch,Straub:CVPR:2017:Cloud,Luo:TIP:2018:trajectoriesTopics,Straub:PAMI:2018:surface,Markovitz:CVPR:2020graph,Uziel:ICCV:2019:BASS,Hayden:2020:Nonparametric}, it has yet to become a mainstream choice, partly due to the lack of efficient large-scale inference tools.
Fortunately, this is starting to change; see, \eg, the highly-effective DPM sampler from~\cite{Dinari:CCGRID:2019:distributed} (a modern and scalable implementation
of the DPM sampler from~\cite{Chang:NIPS:2013:ParallelSamplerDP})
or the scalable streaming DPM inference in~\cite{Dinari:AISTATS:2022:Streaming}. 
Of note, an important alternative to sampling is variational DPM inference~\cite{BNP:Blei:2006:variationalDPMM,BNP:Hoffman:2013:SoVB, BNP:Hughes:2013:MoVB, BNP:Kurihara:NIPS:2007:accVDPMM, BNP:Huynh:ACCL:2016:streamingBNP}. 
A non-Bayesian example of a popular nonparametric method is DBSCAN~\cite{BNP:Ester:1996:DBSCAN} which is density-based and groups together closely-packed points.
While DBSCAN has efficient implementations, it is highly sensitive to its hyperparameters which are hard to tune.

\textbf{Nonparametric Deep Clustering.}
Among the very few examples of deep methods that also find $K$ are~\cite{DC_BNP:Chen:2015:deepbeliefBNP, DC_BNP:Zhao:2019:adapVAE, DC_BNP:Wang:2021:BNP_dnb, DC_BNP:Shah:2018:DCC}.
Some of them use an offline DPM inference for pseudo-labels for fine-tuning a deep belief network~\cite{DC_BNP:Chen:2015:deepbeliefBNP},
or an AE~\cite{DC_BNP:Wang:2021:BNP_dnb} (similarly to the parametric methods in~\cite{Yang:ICML:2017:towards,Caron:2018:ECCV:deepcluster,Caron:2019:ICCV:unsupervised}).
As the methods in~\cite{DC_BNP:Wang:2021:BNP_dnb} and~\cite{DC_BNP:Chen:2015:deepbeliefBNP} rely on slow DPM samplers, they do not scale to large datasets.
AdapVAE~\cite{DC_BNP:Zhao:2019:adapVAE} uses a DPM prior for a VAE.  
In DCC~\cite{DC_BNP:Shah:2018:DCC}, feature learning and clustering are performed simultaneously like in~\cite{DC_BNP:Zhao:2019:adapVAE}; however, instead of ELBO minimization, DCC uses a nearest-neighbor graph to group points that are close in the latent space of an AE.
Our method is empirically more effective than~\cite{DC_BNP:Zhao:2019:adapVAE,DC_BNP:Shah:2018:DCC} 
and also scales much better. 
While not a clustering method per-se, and similarly to~\cite{DC:Mcconville:ICPR:2021:n2d},~\cite{wang2018deep} uses an AE and t-SNE~\cite{van2008visualizing:tSNE} to find $K$. Like~\cite{DC:Mcconville:ICPR:2021:n2d}, however,~\cite{wang2018deep} does not scale.
In~\cite{Dosovitskiy:ICLR:2020:YouOnlyTrainOnce}, a deep net is simultaneously trained on a family of losses instead of a single one. At least in theory, that approach may be adapted to nonparametric clustering but this direction has yet to be explored. 
Both~\cite{Tapaswi:ICCV:2019:ClusteringFaces} and~\cite{Pakman:ICLR:2020:NPC} do not assume a known $K$,
where the former focuses on clustering faces and the latter on generating posterior samples of cluster labels for
any new dataset. 
Unlike our method, however,~\cite{Tapaswi:ICCV:2019:ClusteringFaces,Pakman:ICLR:2020:NPC} are supervised. 
Similarly,~\cite{Avgerinos:IEEE:2020:Iterative} iteratively forms clusters by sequentially examining each sample against the members of existing clusters. The clustering criterion is based on a supervised evaluation net. 
Lastly, while~\cite{DC_BNP:Yang:2019:vsb} relies on a BNP mixture, their method (and code) still uses a fixed $K$.

\section{Preliminaries: DPGMM-based Clustering}\label{Sec:Prelims}
Let $\Xcal=(\bx_i)_{i=1}^N$ denote $N$ data points in $\Rd$.
The clustering task aims to partition $\Xcal$
into $K$ disjoint groups, where $z_i$ is the point-to-cluster assignment,
known as the \emph{cluster label}, 
of $\bx_i$. \emph{Cluster} $k$ consists of all the points labeled as $k$;
\ie, $(\bx_i)_{i:z_i=k}$. 
The number of clusters, $K\triangleq|\set{k:k\in \bz}|$, is thus the number of unique elements in $\bz=(z_i)_{i=1}^N$.

The classical Gaussian Mixture Model (GMM) has a BNP extension: the Dirichlet Process GMM (DPGMM)~\cite{Ferguson:1973:BNP,Antoniak:1974:DPMM}. Informally, the DPGMM (a specific case of the DPM) entertains the notion of a mixture with infinitely-many Gaussians: 
\vspace{-0.5em}
\begin{align}
 p(\bx|(\bmu_k,\bSigma_k,\pi_k)_{k=1}^\infty)
  =    \sum\nolimits_{k=1}^{\infty} \pi_k \Ncal(\bx; \bmu_k, \bSigma_k) \, 
\end{align}
where $\Ncal(\bx;\bmu_k,\bSigma_k)$ is 
a
Gaussian  probability density function (pdf) (of mean $\bmu_k\in\Rd$ and a $d$-by-$d$ covariance matrix $\bSigma_k$) evaluated at $\bx\in \Rd$,
$\pi_k>0$  $\forall k$, and $\sum_{k=1}^\infty \pi_k=1$. 
For a gentle introduction to the DPGMM 
 with a computer-vision audience in mind, see~\cite{Sudderth:PhD:2006:GraphicalModels,Chang:Thesis:2014:Sampling}. 
 Let $\btheta_k=(\bmu_k,\bSigma_k)$ denote the parameters of Gaussian $k$.
Note the distinction between \emph{component} $k$ (namely, the $k$-th Gaussian, 
identified with its parameter, $\btheta_k$)
and cluster $k$.
The components, $\btheta=(\btheta_k)_{k=1}^\infty$, and weights, $\bpi=(\pi_k)_{k=1}^\infty$, are assumed to be drawn (independently) from their own prior distributions:
the weights, $\bpi$, are drawn using the
Griffiths-Engen-McCloskey stick-breaking process (GEM)~\cite{Pitman:Book:2002:Combinatorial} with a concentration parameter $\alpha>0$, 
while the parameters, 
$\tuple{\btheta_k}_{k=1}^\infty$, 
are independent and identically-distributed (\iid) draws 
from their prior
$p(\btheta_k)$, typically a Normal-Inverse Wishart (NIW) distribution.
While there are infinitely-many \emph{components}, note that there are still finitely-many \emph{clusters}
as the \emph{latent random variable} $K$ is bounded above by $N$. 
By possibly renaming cluster indices,  we may assume without loss of generality that $\set{k:k\in \bz}=\set{1,2,\ldots,K}$.
{

The DPGMM is often used in clustering when $K$ is unknown. 
DPGMM inference methods typically seek
to  find $\bz=(z_i)_{i=1}^N$ (which implies $K$) and $(\btheta_k,\pi_k)_{k=1}^K$.
As explained in our supplementary material (\textbf{Supmat}),  
the inferred value of $K$ is affected by the following factors:
$\Xcal$, $\alpha$, and the NIW hyperparameters. 
Our method (\autoref{Sec:Method}) is inspired in part by Chang and Fisher III's DPM sampler~\cite{Chang:NIPS:2013:ParallelSamplerDP}
which consists of a split/merge framework~\cite{BNP:Jain:2004} (which we adopt)
and a restricted sampler (which is less relevant to our work). 

The split/merge framework augments the latent variables, $(\btheta_k)_{k=1}^\infty$, $\bpi$, and $(z_i)_{i=1}^N$, with auxiliary variables. To each $z_i$, an additional \emph{subcluster label}, $\widetilde{z}_i\in\set{1,2}$, is added. 
To each $\btheta_k$, two subcomponents are added,
$\widetilde{\btheta}_{k,1}$, $\widetilde{\btheta}_{k,2}$,
with nonnegative weights 
$\widetilde{\bpi}_k=(\widetilde{\pi}_{k,j})_{j\in\set{1,2}}$ (where $\widetilde{\pi}_{k,1}+\widetilde{\pi}_{k,2}=1$), forming a 2-component GMM. 
Next, splits and merges allow changing $K$ 
via the Metropolis-Hastings framework~\cite{Hastings:1970:MH}. That is, during the inference, every certain amount of iterations 
the split of cluster $k$ into its subclusters is proposed.
That split is accepted with probability $\min(1,H_\mathrm{s})$ where
\begin{align}
\hspace{-.2cm}        H_\mathrm{s} = \frac{\alpha \Gamma(N_{k,1}) f_\bx(\Xcal_{k,1};\lambda)  \Gamma(N_{k,2}) f_\bx(\Xcal_{k,2};\lambda)}{\Gamma(N_k) f_\bx(\Xcal_k;\lambda)} 
        \label{eqn:HastingRatioSplit}
\end{align}
is the Hastings ratio, $\Gamma$ is the Gamma function, $\Xcal_k=(\bx_i)_{i:z_i=k}$ stands for the points in cluster $k$,
 $N_k = |\Xcal_k|$, $\Xcal_{k,j}=(\bx_i)_{i:(z_i,\widetilde{z}_i)=(k,j)}$
denotes the points in subcluster $j$ ($j\in\set{1,2}$), $N_{k,j} = |\Xcal_{k,j}|$, and $f_\bx(\cdot;\lambda)$
is the \emph{marginal} likelihood where $\lambda$
represents the NIW hyperparameters. See our \textbf{Supmat} for more details. 
Upon a split proposal
acceptance, each of the newly-born clusters is augmented with two subclusters. This ratio, $ H_\mathrm{s}$, can be interpreted as comparing the marginal likelihood of the data under the two subclusters with its marginal likelihood under the cluster. Merge proposals are handled similarly (see \textbf{Supmat}).
\begin{figure*}[!t]
\hspace{.4cm}
\includegraphics[scale=.73,trim={3.5mm 182mm 30mm 0mm},clip]{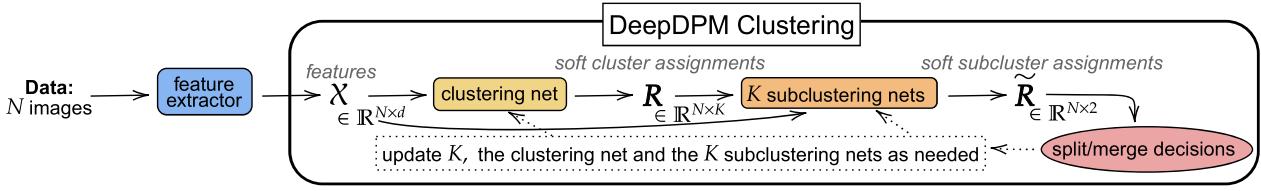}
\WHITE{--------} \vspace{-1cm} \newline
\captionsetup{justification=justified, singlelinecheck=false}
\caption{DeepDPM's pipeline: given features $\Xcal$, the clustering net outputs cluster assignments, $\bR$, while the subclustering nets generate subcluster assignments, $\widetilde{\bR}$. Upon the acceptance of split/merge proposals, 
all those nets are updated during the learning.}
\label{method_figs:clustering_pipeline}
\vspace{-.2cm}
\end{figure*}

\section{The Proposed Method: DeepDPM}\label{Sec:Method}
DeepDPM can be viewed as a DPM inference algorithm.
Inspired by~\cite{Chang:NIPS:2013:ParallelSamplerDP}, we use splits and merges to change $K$ where for every cluster we maintain a subcluster pair. For a nominal value of $K$, 
rather than resorting to sampling as in~\cite{Chang:NIPS:2013:ParallelSamplerDP}, 
we use a deep net trained by a novel amortized inference for EM~\cite{Dempster:JRSS:1977:EM} in a mixture model. 
DeepDPM has two main parts. The first is a clustering net,
while the second consists of $K$ subclustering nets (one for each cluster $k$, $k\in\set{1,\ldots,K}$). 
In~\autoref{methods:clustering_net} we describe how DeepDPM operates given a nominal value of $K$ and in~\autoref{method:split_merge} how $K$ is changed and how our architecture adapts accordingly. 
We discuss the amortized-inference aspects of our approach in~\autoref{Sec:Method:Amortized}, our weak prior in~\autoref{Sec:Method:Prior}, and how DeepDPM may be combined with feature learning in~\autoref{method:feature_extraction}.  
\autoref{method_figs:clustering_pipeline} depicts the overall pipeline.
\subsection{DeepDPM Under a Fixed $K$}\label{methods:clustering_net}
We start by describing DeepDPM's forward pass during the training.  
Given a current value of $K$, the data is first passed to the clustering net, $f_\mathrm{cl}$,  which generates,
for each data point $\bx_i$, $K$ soft cluster assignments: 
\begin{align}
	f_\mathrm{cl}(\Xcal) = \bR=(\br_i)_{i=1}^N\qquad \br_i=(r_{i,k})_{k=1}^K
\end{align}
where $r_{i,k}\in[0,1]$ is the soft assignment of $\bx_i$ to cluster $k$
(also called the responsibility of cluster $k$ to data point $\bx_i$) 
and $\sum_{k=1}^K r_{i,k} = 1$.
From $(\br_i)_{i=1}^N$ we compute the hard assignments $\bz = (z_i)_{i=1}^N$ by $z_i = \argmax{k}r_{i,k}$. Next, each subclustering net, $f_{\mathrm{sub}}^{k}$ (where $k\in\set{1,\ldots,K}$), is fed with the data (hard-) assigned to its respective cluster (\ie, $f_{\mathrm{sub}}^{k}$ is fed with $\Xcal_k = (\bx_i)_{i:z_i = k}$) and generates soft subcluster assignments:
\begin{align}
	f_{\mathrm{sub}}^{k}(\Xcal_k) = \widetilde{\bR}_k=(\widetilde{\br}_i)_{i:z_i=k} 
	\qquad \widetilde{\br}_i=(\widetilde{r}_{i,j})_{j=1}^2
\end{align}
where  $\widetilde{r}_{i, j} \in [0,1]$ is the soft assignment of $\bx_i$ to subcluster $j$ ($j\in\set{1,2}$), and 
$\widetilde{r}_{i,1}  +\widetilde{r}_{i,2} = 1\, \forall k\in\set{1,\ldots,K}$. As detailed in~\autoref{method:split_merge}, the subclusters learned by $(f_{\mathrm{sub}}^k)_{k=1}^K$ are used in split proposals.
Each of the $K+1$ nets ($f_\mathrm{cl}$ and $(f_\mathrm{sub}^k)_{k=1}^K$) is a simple multilayer perceptron with a single hidden layer. The last layer of $f_\mathrm{cl}$ has $K$ neurons while the last layer of each $f_\mathrm{sub}^k$ has two.  

We now introduce a new loss motivated by EM in the Bayesian GMM
(though the idea, in fact, also holds in the non-Bayesian GMM case, as well 
as for EM in parametric mixtures with non-Gaussian components).  
Concretely, in each epoch, our clustering net is optimized to generate soft assignments that would resemble those obtained by an E step of the EM-GMM algorithm (recall that the E steps of the Bayesian and non-Bayesian EM-GMM coincide). 
Each E step is followed by a standard M step in a Bayesian GMM, except that the soft assignments used in the Maximum-a-Posterior (MAP) estimates
are those produced by our clustering net. We now provide the details. 
For each $\bx_i$ and each $k\in\set{1,\ldots,K}$ we compute the (standard) E-step probabilities, 
$
\br_i^{\mathrm{E}}=
    (r_{i,k}^{\mathrm{E}})_{k=1}^K$, 
    where
\begin{align}
\hspace{-1mm}
    r_{i,k}^{\mathrm{E}} = \frac{\pi_k  \Ncal(\bx_i; \bmu_k, \bSigma_k)}{\sum_{k'=1}^K \pi_{k'} \Ncal(\bx_i; \bmu_{k'}, \bSigma_{k'})} \quad  k\in\set{1,\ldots,K}
    \label{Eqn:Estep}
\end{align}
is computed using $(\pi_k,\bmu_k,\bSigma_k)_{k=1}^K$ from the previous epoch. 
Note that $\sum_{k=1}^K r_{i,k}^{\mathrm{E}}=1$. 
We then encourage $f_\mathrm{cl}$ to generate similar soft assignments 
using the following new loss: 
\begin{align}
	\Lcal_\mathrm{cl} =
	\sum\nolimits_{i=1}^N \mathrm{KL}(\br_i\|\br_i^{\mathrm{E}})
\end{align}
where KL is the Kullback-Leibler divergence. 
Next, after every epoch we perform a Bayesian M step but with a twist. 
Recall that in this step, one uses the weighted versions of the MAP estimates 
of $(\pi_k,\bmu_k,\bSigma_k)_{k=1}^K$ (computed using standard formulas; see~\textbf{Supmat}) where the weights are the $r_{i,k}^{\mathrm{E}}$ values (\EQN\eqref{Eqn:Estep}). We apply the same formulas but instead of $r_{i,k}^{\mathrm{E}}$ we use the $r_{i,k}$ values (\ie, the output of $f_\mathrm{cl}$). 
Note that unlike methods (\eg $K$-means or SCAN) that enforce/assume uniformity of the weights, 
our inferred cluster weights, $(\pi_k)_{k=1}^K$, are allowed to deviate  
from uniformity. 

In principle, for $(f_{\mathrm{sub}}^k)_{k=1}^K$  we could have used a loss similar to
$\Lcal_{\mathrm{cl}}$. However, 
here we prefer an isotropic loss:  
\begin{align}
\hspace{-.2cm}
    \Lcal_{\mathrm{sub}} = \sum\nolimits_{k=1}^K\sum\nolimits_{i=1}^{N_k}\sum\nolimits_{j=1}^2
    \widetilde{r}_{i,j}\ellTwoNorm{\bx_i - \widetilde{\bmu}_{k,j}}^2 
\end{align}
where $N_k=|\Xcal_k|$ and $\widetilde{\bmu}_{k,j}$ is the mean of subcluster $j$ of cluster $k$,
computed after every epoch, alongside with subcluster weights and covariances, 
using weighted MAP estimates similarly to the clusters' case (see \textbf{Supmat}).
This loss is more efficient than the KL loss while the latter (only in the subcluster case) did not yield improvement. 
The iterative process described above needs to be initialized. We do this using $K$-means (using some initial value of $K$ for the clustering and $K=2$ for the subclustering).
 \emph{DeepDPM is fairly robust to the initial $K$ so the latter can be chosen arbitrarily} (see, \eg,~\autoref{Sec:Ablation}).

\subsection{Changing $K$ via Splits and Merges}\label{method:split_merge}
During training, we use splits and merges to change $K$ (as in~\cite{Chang:NIPS:2013:ParallelSamplerDP}). 
Every few epochs, we propose either splits or merges. Since $K$ is changing, the architecture, and more specifically, the last layer of the clustering net and the number of subclustering nets, must change too. 
Of note, the splits/merges facilitate not only changing the value of $K$ but also  \emph{large moves}, escaping many poor local optima~\cite{Chang:NIPS:2013:ParallelSamplerDP}. 

\textbf{Splits.} In every split step, we propose to split each of the clusters into its two subclusters. A split proposal is accepted stochastically (as in~\cite{Chang:NIPS:2013:ParallelSamplerDP}) with probability $\min(1,H_\mathrm{s})$; see \EQN\eqref{eqn:HastingRatioSplit}.
To accommodate for the increase in $K$, if a split proposal is accepted for cluster $k$, the $k$-th unit in the last layer of the clustering net, together with the weights connecting it to the previous hidden layer, is duplicated, and
we initialize the parameters of the two new clusters 
using the parameters
learned via the subcluster nets: 
\begin{align}
	&\bmu_{k_1} \gets \widetilde{\bmu}_{k,1}\,,\,
	&\bSigma_{k_1} \gets  \widetilde{\bSigma}_{k,1}\,,\quad
	&\pi_{k_1} \gets  \pi_{k} \times \widetilde{\bpi}_{k,1} \nonumber \\	
	&\bmu_{k_2} \gets  \widetilde{\bmu}_{k,2}\,,\,
    &\bSigma_{k_2} \gets  \widetilde{\bSigma}_{k,2}\,,\quad 	
    &\pi_{k_2} \gets  \pi_{k} \times \widetilde{\bpi}_{k,2}\,  
\end{align}
where $k_1$ and $k_2$ denote the indices of the new clusters. We then also add, to each new cluster, a new subclustering net (dynamically allocating the memory). 

\textbf{Merges.} When considering merges
we must ensure we never 
simultaneously accept the proposals of, \eg, merging clusters $k_1$ and $k_2$ and the merging of clusters $k_2$ and $k_3$, 
thereby mistakenly merging \emph{three} clusters together. Thus, unlike split proposals which are done in parallel, 
not all possible merges can be considered simultaneously.
To avoid sequentially considering all possible merges, we consider (sequentially) the merges of each cluster with only its
3 nearest neighbors. 
Merge proposals are accepted/rejected using a Hastings ratio, $H_{\mathrm{m}}=1/H_{\mathrm{s}}$ (as in~\cite{Chang:NIPS:2013:ParallelSamplerDP}).
If a proposal is accepted, the two clusters are merged and a new subclustering network is initialized. Technically, one of the merged clusters' last layer's units, together with the net's weights connecting it to the previous hidden layer, is removed from $f_{\mathrm{cl}}$, and the parameters and the weight of the newly-born cluster are initialized using the weighted MAP estimates.

\subsection{Amortized EM Inference}\label{Sec:Method:Amortized}
Suppose we turn off splits/merges and use the Ground Truth (GT) $K$. 
Seemingly, this reduces each training epoch to 
mimicking a single EM iteration. Remarkably, however, and as shown in~\autoref{Sec:Results}, even then our method
still yields results that are \emph{usually better} than the standard EM.
We hypothesize that this stems from the fact that we \emph{amortize} the EM inference;
by the virtue of the smoothness of the function learned by the deep net, we improve
the prediction for the points in not only the current batch but also other batches. Moreover, the smoothness also serves as an inductive bias such that
points which are close in the observation space should have similar
labels. 

In principle, instead of using our variational loss we could have also used the GMM  negative log likelihood (or log posterior). However, empirically that led to unstable optimization and/or poor results. Moreover, 
basing our loss on matching soft labels rather than likelihood/posterior elegantly makes the method 
more general:  $f_\mathrm{cl}$ and $\Lcal_\mathrm{cl}$ can be used as they are for any component type, not just Gaussians.

\begin{table*}[!t]
    \centering
\begin{tabular}{@{} >{\arraybackslash}p{0.14\linewidth} c c c c c c c c c @{}}
\toprule
 & NMI & ARI & ACC & NMI & ARI & ACC & NMI & ARI & ACC\\
\midrule
& \multicolumn{3}{c}{MNIST~\cite{Deng:2012:Mnist}} & \multicolumn{3}{c}{USPS~\cite{uspsdataset}} & \multicolumn{3}{c}{Fashion-MNIST~\cite{Xiao:2017:Fashion-MNIST}}\\
\midrule
$K$-means$^p$ & .90$\pm$ .02 & .84$\pm$ .05 & .85$\pm$.06 &
.86$\pm$.01 & .79$\pm$.05 & .80$\pm$.06 &
.67$\pm$.01& .50$\pm$.03 & .60$\pm$.04 
\\
GMM$^p$ & \textbf{.94$\pm$.00} & \textbf{.95$\pm$.00} & \textbf{.98$\pm$.00} &
.86$\pm$.02 & .79$\pm$.05 & .81$\pm$.06 &
.66$\pm$.01& .49$\pm$.02 & .58$\pm$.03
\\
\hline
DBSCAN &  .92$\pm$0 &.86$\pm$0  & .89$\pm$0 &
.72$\pm$0 & .46$\pm$0 & .57$\pm$0 & 
.63$\pm$0 & -.32$\pm$0 &  .39$\pm$0\\
DPM Sampler  & .92$\pm$.01 & .91$\pm$.04 & .93$\pm$.05 &
.87$\pm$.01 & .82$\pm$.02 & .83$\pm$.03 &
.67$\pm$.01 & .49$\pm$ .02 & .59$\pm$.03\\
moVB & .93$\pm$.00 & .94$\pm$.00& .97$\pm$.00
&.87$\pm$.02 &\textbf{.86$\pm$.04} & \textbf{.90$\pm$.04} &
.66$\pm$.02 & .47$\pm$.03 & .55$\pm$.03\\
DeepDPM (Ours) & \textbf{.94$\pm$.00} & \textbf{.95$\pm$.00} & \textbf{.98$\pm$.00} 
& \textbf{.88$\pm$.00} & \textbf{.86$\pm$.01} &.89$\pm$.2 &
\textbf{.68$\pm$.01} & \textbf{.51$\pm$.02} &\textbf{.62$\pm$.03}
\\
\bottomrule
\toprule

& \multicolumn{3}{c}{MNIST$^{imb}$} & \multicolumn{3}{c}{USPS$^{imb}$} & \multicolumn{3}{c}{Fashion-MNIST$^{imb}$}\\
\midrule
$K$-means$^p$ & .89$\pm$ .03 & .84$\pm$ .06 & .83$\pm$.06 &
.82$\pm$.02 & .71$\pm$.05 & .71$\pm$.05 &
.62$\pm$.01& .46$\pm$.02 & .56$\pm$.03
\\
GMM$^p$ & .94$\pm$.02 & .95$\pm$.03 & .96$\pm$.04 &
.83$\pm$.01 & .74$\pm$.05 & .76$\pm$.05 &
.62$\pm$.01& .46$\pm$.02 & .57$\pm$.03
\\
\hline
DBSCAN &  .93$\pm$0 &.92$\pm$0  & .94$\pm$0 &
.84$\pm$0 & .79$\pm$0 & .80$\pm$0 & 
.62$\pm$0 & .35$\pm$0 &  .46$\pm$0\\
DPM Sampler & .93$\pm$.01 & .94$\pm$.02& .96$\pm$.02 &
.89$\pm$.02 & .89$\pm$.06 & .91$\pm$.04 &
\textbf{.66$\pm$.01} & \textbf{.50$\pm$ .01} & \textbf{.61$\pm$.01}\\
moVB & .94$\pm$.00 & .95$\pm$.00& .96$\pm$.00
&.88$\pm$.01 &.89$\pm$.02 & .91$\pm$.02 &
.63$\pm$.01 & .44$\pm$.02 & .53$\pm$.02\\
DeepDPM (Ours) & \textbf{.95$\pm$.01} & \textbf{.97$\pm$.01} & \textbf{.98$\pm$.01} 
& \textbf{.90$\pm$.00} & \textbf{.92$\pm$.00} &\textbf{.94$\pm$.00} &
.65$\pm$.00 & \textbf{.50$\pm$.00} &\textbf{.61$\pm$.00}
\\
\bottomrule
    \end{tabular}
\caption[]{Comparing the mean results ($\pm$std.~dev.) of DeepDPM with classical clustering methods. The results are the mean of 10 independent runs. Methods marked with $^p$ are parametric (require $K$). Datasets marked with $^{imb}$ are imbalanced ones.}
\label{results:umap_embedded}
\end{table*}

\subsection{A Weak Prior: Letting the Data Speak for Itself}\label{Sec:Method:Prior}
Recall that the inferred $K$ depends on $\Xcal$, $\alpha$, and the NIW hyperparameters. 
We intentionally choose the prior to be \emph{very} weak. Meaning, we choose $\alpha$
as well as the so-called pseudocounts (two of the NIW hyperparameters) to be very low numbers, dwarfed by $N$, the number of points (see \textbf{Supmat} for details). Thus, we let the data, $\Xcal$, to be the most dominant factor in determining $K$. The weak prior also means that the Bayesian EM-GMM nearly coincides with the non-Bayesian EM-GMM but still helps in the presence of a degenerate sample covariance or very small clusters.

\subsection{Feature extraction}\label{method:feature_extraction}
To show the effectiveness of our clustering method, we used two types of feature-extraction paradigms: an end-to-end approach, where features and clustering are learned jointly (using alternate optimization), and a two-step approach in which features are learned once (before the clustering) and then held fixed.
For the two-step approach, we follow SCAN~\cite{Gansbeke:ECCV:2020:SCAN} and use MoCo~\cite{DC:Chen:2020:MoCO} for (unsupervised) feature extraction. For more details, as well as the scheme we use for an end-to-end feature extraction, see \textbf{Supmat}.
\section{Results}\label{Sec:Results}
In this section we evaluate DeepDPM and compare it with several key methods on popular image and text datasets at varying scales.
In our evaluations we use three common metrics: clustering accuracy (ACC); Normalized Mutual Information (NMI);  Adjusted Rand Index (ARI). The higher the better in all three, and they can accommodate for differences between the inferred $K$ and its GT value.
See \textbf{Supmat} for more details on the experimental setup and 
the values of the hyperparameters that we used. 
Due to space limits, we omit here the NMI and ARI values in several comparisons but these appear in the \textbf{Supmat}.
We round the results to 2 decimal places so, \eg, a standard deviation (std.~dev.) of .00 may still represent a positive (albeit small) number.
\begin{table}
    \centering
\begin{tabular}{@{}p{2.58cm} >{\centering\arraybackslash}p{0.16\linewidth} >{\centering\arraybackslash}p{0.144\linewidth} >{\centering\arraybackslash}p{0.28\linewidth} @{}}
\toprule
Method & \multicolumn{3}{c}{Inferred $K$}\\ 
\midrule
& MNIST & USPS & Fashion-MNIST\\
\midrule
DBSCAN & 9.0$\pm$0.00 & 6.0$\pm$0.00 & 4.0$\pm$0.00\\
DPM Sampler & 11.3$\pm$0.82 & 8.5$\pm$0.85 & 12.4$\pm$0.97\\
moVB & 14$\pm$1.00 & 11.2$\pm$1.08 & 16.9$\pm$2.30\\
DeepDPM (Ours) & \textbf{10$\pm$0.00} & \textbf{9.2$\pm$0.42} & \textbf{10.2$\pm$0.79}\\
\bottomrule
    \end{tabular}
\caption[]{Comparing the mean inferred value ($\pm$std.~dev.) for $K$ of 10 runs among nonparametric methods. GT $K=10$.}
\label{results:umap_embedded_k}
\end{table}				  
 
\begin{table*}[h!]
\centering
    \begin{tabular}{@{}p{.182\linewidth} >{\centering\arraybackslash}p{0.065\linewidth}>{\centering\arraybackslash}p{0.065\linewidth}>{\centering\arraybackslash}p{0.065\linewidth}>{\centering\arraybackslash}p{0.065\linewidth}>{\centering\arraybackslash}p{0.065\linewidth}>{\centering\arraybackslash}p{0.065\linewidth}>{\centering\arraybackslash}p{0.065\linewidth}>{\centering\arraybackslash}p{0.065\linewidth}>{\centering\arraybackslash}p{0.065\linewidth}@{}}
    \toprule
    & \multicolumn{3}{c}{MNIST~\cite{Deng:2012:Mnist}} & \multicolumn{3}{c}{STL-10~\cite{Coates:AISTATS:2011:STL-10}} & \multicolumn{3}{c}{Reuters10k~\cite{Lewis:2004:rcv1}} \\
    \midrule
    Method & NMI & ARI & ACC & NMI & ARI & ACC  & NMI & ARI & ACC \\ \midrule
    \hline
    AdapVAE$\dagger$~\cite{DC_BNP:Zhao:2019:adapVAE} \textit{avg} & .86$\pm$1.02 & .84$\pm$2.35 & N/A & .75$\pm$0.53 & \textbf{.71$\pm$0.81} & N/A & .45$\pm$1.79 & .43$\pm$5.73 & N/A\\
    DCC$\dagger$~\cite{DC_BNP:Shah:2018:DCC} best & .912 & N/A & .96 & N/A & N/A & N/A & .59 & N/A & .60\\
    DCC$\ddagger$~\cite{DC_BNP:Shah:2018:DCC} \textit{avg} & .90$\pm$.02 & .89$\pm$.07 & .91$\pm$.07 & .22$\pm$.00 & .01$\pm$.00 & .04$\pm$.00 & .25$\pm$.00 & .00$\pm$.00 & .00$\pm$.00\\
    \midrule
    DeepDPM (ours) \textit{avg} & .90$\pm$.01 & .91$\pm$.02 & .93$\pm$.03 & .78$\pm$.004 & .70$\pm$.01 & .84$\pm$.01 & .61$\pm$.00 & .64$\pm$.01 & .83$\pm$.00\\
    DeepDPM (ours) best & \textbf{.92} & .\textbf{93} & \textbf{.96} &\textbf{.79} & \textbf{.71} & \textbf{.85} & \textbf{.61} & \textbf{.64} & \textbf{.83}\\
 
    \bottomrule
    \end{tabular}
\caption[Comparing to Deep non-parametric methods]{Comparing deep nonparametric methods. $\dagger$: reported in the papers. $\ddagger$: obtained using their code. \textit{avg}: mean ($\pm$std.~dev.) of 5 runs.
}
\label{results:end2end}
\end{table*}

\textbf{Comparing with Classical Methods.}
We compared DeepDPM with classical parametric methods ($K$-means; GMM) and nonparametric ones (DBSCAN~\cite{BNP:Ester:1996:DBSCAN}, moVB~\cite{BNP:Hughes:2013:MoVB}; the SOTA DPM sampler from~\cite{Dinari:CCGRID:2019:distributed}).
For feature extraction, we performed the process suggested in~\cite{DC:Mcconville:ICPR:2021:n2d}. 
We performed the evaluation on the MNIST, USPS, and Fashion-MNIST datasets, as
well their imbalanced versions (the latter are defined in the \textbf{Supmat}).
All the methods used the same (and fixed) data embeddings as input, \emph{and the parametric ones were given the GT $K$, given them an unfair advantage}.
\autoref{results:umap_embedded} shows that DeepDPM almost uniformly dominates across all datasets and metrics,
and its performance gain only increases in the imbalanced cases.
It is also observable that, compared with the parametric methods, the nonparametric ones (ours included) are less affected by the imbalance. Moreover,~\autoref{results:umap_embedded_k} shows that among the nonparametric methods, DeepDPM's inferred $K$ is the closest to the GT $K$ (see \textbf{Supmat}
for similar results in the imbalanced case).
\begin{table}[h!]
\centering
    \begin{tabular}{@{}l ccc@{}}
    \toprule
    Method & NMI & ARI & ACC\\
    \midrule
    & \multicolumn{3}{c }{ImageNet-50: Balanced} \\
    \midrule
    DBSCAN & .52$\pm$.00 & .09$\pm$.00 & .24$\pm$.00 \\
    moVB & .70$\pm$.01 & .38$\pm$.01 & .55$\pm$.02\\
    DPM Sampler & .72$\pm$.00 & .43$\pm$.01 & .57$\pm$.01 \\
    DeepDPM (ours) & .75$\pm$.00 & .49$\pm$.01 & .64$\pm$.00 \\
    DeepDPM (ours)$\ast$ & \textbf{.77$\pm$.00} & \textbf{.54$\pm$.01} & \textbf{.66$\pm$.01} \\
    \midrule
    \midrule
    & \multicolumn{3}{c }{ImageNet-50: Imbalanced}\\
    \midrule
    DBSCAN & .33$\pm$.00 & .04$\pm$.00 & .24$\pm$.00\\
    moVB & .68$\pm$.01 & .44$\pm$.03 & .52$\pm$.03\\
    DPM Sampler & .70$\pm$.00 & .40$\pm$.01 & .51$\pm$.00\\
    DeepDPM (ours) & .74$\pm$.01 & .48$\pm$.02 & .58$\pm$.01 \\
    DeepDPM (ours)$\ast$ & \textbf{.75$\pm$.00} & \textbf{.51$\pm$.01} & \textbf{.60$\pm$.01}\\
    \bottomrule
    \end{tabular}
\caption[]{Comparison of nonparametric methods on ImageNet-50 and its imbalanced version. $\ast$ marks results with AE alternation.}
\label{results:non_param}
\vspace{-2mm}
\end{table}

\textbf{Comparing with Deep Nonparametric Methods.}
As there exist very few deep nonparametric methods, and some of them reported results only on \emph{extremely-small} toy datasets~\cite{DC_BNP:Wang:2021:BNP_dnb, DC_BNP:Chen:2015:deepbeliefBNP} (\eg, one of them stated they could not process even MNIST's train dataset as it was too large for them), 
we compared DeepDPM with DCC~\cite{DC_BNP:Shah:2018:DCC} and AdapVAE~\cite{DC_BNP:Zhao:2019:adapVAE}, the only unsupervised deep nonparametric methods that can at least handle the MNIST~\cite{Deng:2012:Mnist}, USPS~\cite{uspsdataset}, and STL-10~\cite{Coates:AISTATS:2011:STL-10} datasets.  As both those methods jointly learn features and clustering, and to show the flexible nature of DeepDPM, we demonstrate its integration with two feature-extraction techniques (described in~\autoref{method:feature_extraction}): an end-to-end pipeline (for MNIST and REUTERS-10k~\cite{Lewis:2004:rcv1}) and a two-step approach using features pretrained by MoCo~\cite{DC:Chen:2020:MoCO} (for STL-10).
Unfortunately, we could not run AdapVAE's published code, and thus resort to including the results reported by them. For DCC, using their code we could reproduce their results only on MNIST, so we compare with both the results we managed to obtain using their code and the ones reported by them. 
Due to these reproducibility issues, we could compare with those methods 
only on the original (\ie, balanced) datasets. \autoref{results:end2end} shows that  DeepDPM outperforms both DCC and AdapVAE.
Note we could not find other unsupervised deep nonparametric methods (let alone with available code) that scale to even these fairly-small datasets.

\begin{figure*}[!t]
\centering
{\includegraphics[width=4.1cm]{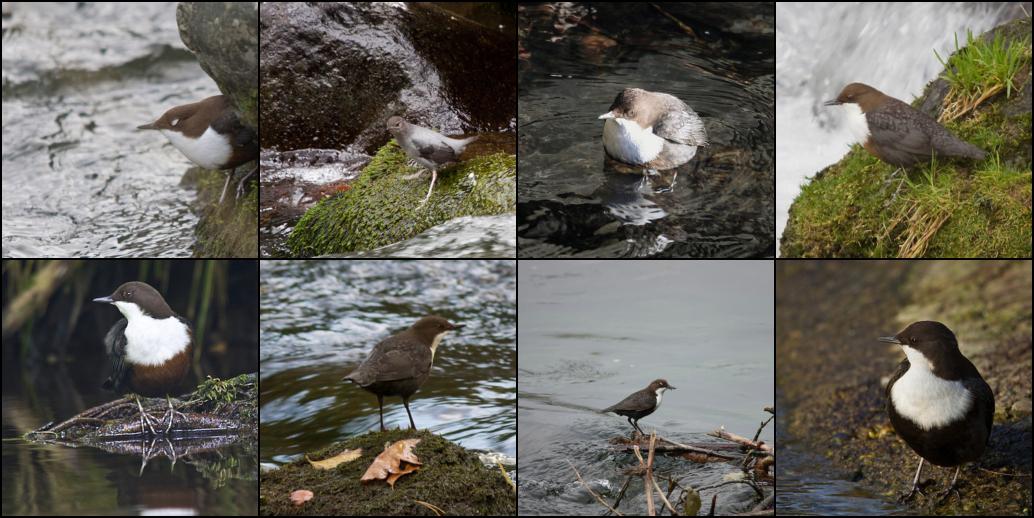}}\hfil
{\includegraphics[width=4.1cm]{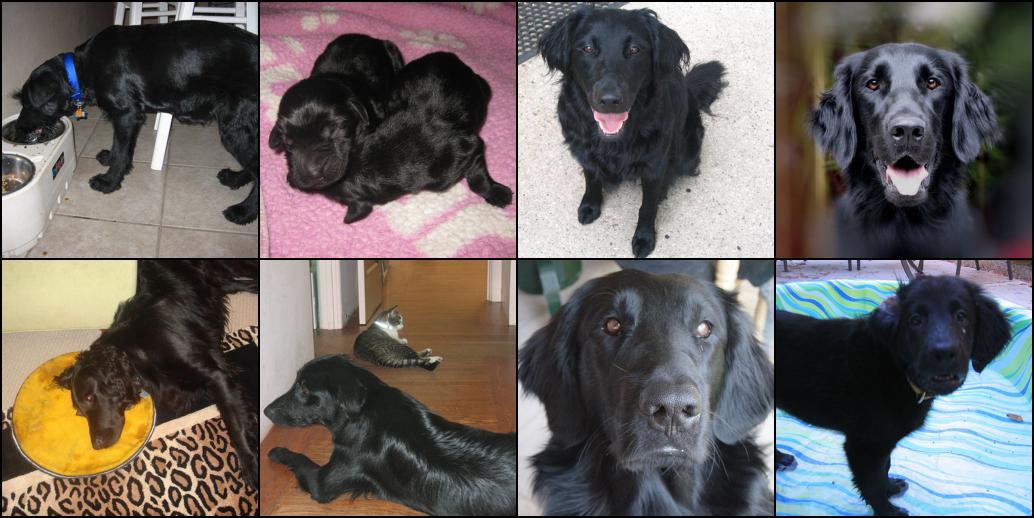}}\hfil
{\includegraphics[width=4.1cm]{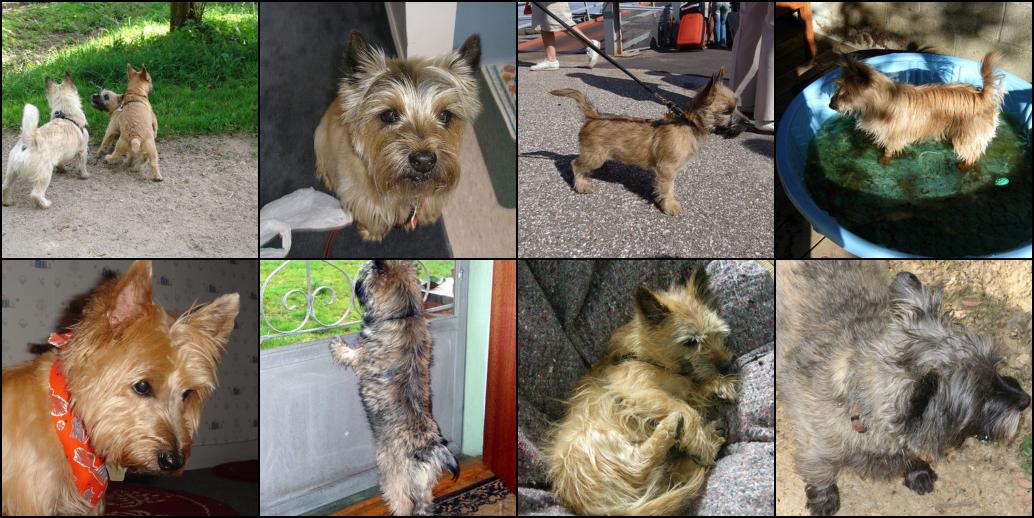}}\hfil
{\includegraphics[width=4.1cm]{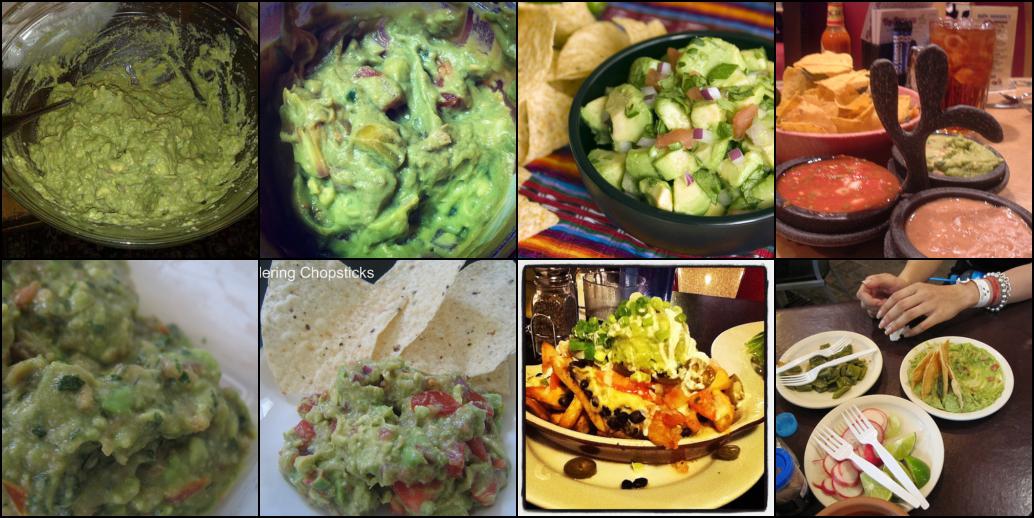}}\hfil
 \vspace{-.2cm}
\caption{Examples of ImageNet images clustered together by DeepDPM. Each panel stands for a different cluster.}\label{Fig:imgs_clustered_together}
\end{figure*}

\textbf{Clustering the Entire ImageNet Dataset.} 
On ImageNet, we obtained the following results: 
ACC: 0.25, NMI: 0.65, ARI: 0.14.
Our method was initialized with $K = 200$ and converged into 707 clusters (GT=1000).
These are  first results on ImageNet reported for deep nonparametric clustering. 
\autoref{Fig:imgs_clustered_together} shows examples of images clustered together.
	
\begin{table}[h!]
\centering
    \begin{tabular}{@{}>{\arraybackslash}p{0.4\linewidth} >{\centering\arraybackslash}p{0.23\linewidth}>{\centering\arraybackslash}p{0.23\linewidth}@{}}
    \toprule
    Method & Final/best $K$: balanced & Final/best $K$: imbalanced\\
    \midrule
    \midrule
    $K$-means$^{p}$ & 40 & 20\\
    DCN++$^{p}$ & 60 & 40 \\
    SCAN$^{p}$ & 70 & 40\\
    \midrule
    DBSCAN & 16 & 13 \\
    moVB & 46.2$\pm$1.3 & \textbf{46.4$\pm$1.1} \\
    DPM Sampler & 72.0$\pm$2.6 & 70.3$\pm$4.6\\
    DeepDPM (ours) & \textbf{52.0$\pm$1.0} & 43.67$\pm$1.2 \\
    DeepDPM (ours)$\ast$ & 55.3$\pm$1.5 & 46.3$\pm$2.5\\
    \bottomrule
    \end{tabular}
\caption[]{Comparing the mean ($\pm$std.~dev.) value for $K$ found on ImageNet-50 of 3 runs. For the parametric methods (marked with $ ^{p}$) we use the $K$ value with the best silhouette score. $\ast$ marks results obtained with AE alternation.}
\label{results:param_non_param}
\vspace{-1mm}%
\end{table}

\subsection{The Value of Deep Nonparametric Methods}
\textbf{When Parametric Methods Break.} 
We study the effect of not knowing $K$ on parametric methods, with and without class imbalance.
We evaluate each method with a wide range of different $K$ values on ImageNet-50. The latter, curated in~\cite{Gansbeke:ECCV:2020:SCAN},
consists of 50 randomly-selected classes of ImageNet~\cite{Deng:CVPR:2009:imagenet}. To generate an imbalanced version of it, we sampled a normalized \emph{nonuniform} histogram from a uniform distribution over the $50$-dimensional probability simplex
(\ie, all histograms were equally probable) and then sampled examples from the 50 classes in proportions according to that nonuniform histogram. We compared with
3 parametric methods: 1) $K$-means; 2) the SOTA method SCAN~\cite{Gansbeke:ECCV:2020:SCAN}; 3) an improved version of DCN~\cite{Yang:ICML:2017:towards}, self-coined DCN++, where instead of training an AE on the raw data, we trained it on top of the embeddings SCAN uses (MoCo~\cite{DC:Chen:2020:MoCO}) where, following~\cite{Gansbeke:ECCV:2020:SCAN}, we froze those embeddings during training. For DeepDPM, we used the same features.

Since SCAN requires large amounts of memory (\eg, we could only run it on 2 RTX-3090 GPU cards with 24GB memory each, compared with DeepDPM for which a single RTX-2080 (or even GTX-1080) with 8GB sufficed), and due to resource constraints, we were limited in how many $K$ values we could run SCAN with and in the number of times each experiment could run (this high computational cost is one of the problems with model selection in parametric methods). Thus,  we collected the results of the parametric methods with $K$ values ranging from 5 to 350.
{For both the balanced and imbalanced cases, we initialized DeepDPM with $K=10$.}
\begin{table}[t]
\centering
    \begin{tabular}{@{}m{3.755cm} m{1.15cm}m{1.29cm}m{1.29cm}@{}}
    \toprule
 & \multicolumn{3}{c}{ACC}\\
    \midrule
    & $K_\mathrm{init}$=3 & $K_\mathrm{init}$=10 &$K_\mathrm{init}$=30\\
    \midrule
    \hline
    No splits/merges & .29$\pm$.01 & .59$\pm$.03 & .46$\pm$.01\\
    No splits & .29$\pm$.01 & .59$\pm$.02 & .45$\pm$.03\\
    No merges & .46$\pm$.00 & .58$\pm$.01 & .47$\pm$.01 \\
    2-means instead of $f_{\mathrm{sub}}$
    & .61$\pm$.00 & .59$\pm$.02 & .56$\pm$.02 \\
    No priors in the $M$ step
    & .58$\pm$.01 & .57$\pm$.02 & .58$\pm$.01 \\
    Isotropic loss instead of $\Lcal_{\mathrm{cl}}$ & .58$\pm$.00 & .58$\pm$.00 & .58$\pm$.02 \\
	DeepDPM (full method) & \textbf{.62$\pm$.03} & \textbf{.61$\pm$.00} & \textbf{.62$\pm$.01}\\  
    \bottomrule
    \end{tabular}
\caption[Ablations]{DeepDPM's performance under different ablations.}
\label{results:DPM_ablations}
\end{table}

\autoref{results_figs:balanced_diff_K} summarizes the ACC results (see \textbf{Supmat} for ARI/NMI). 
As the $K$ value used by the parametric methods diverges from the GT (\ie, $K=50$), their results deteriorate.
Unsurprisingly, when using the GT $K$, or sufficiently close to it, the parametric methods outperform our nonparametric one, confirming our claim that having a good estimate of $K$ is important for good clustering. 
\autoref{balanced}, however,  shows that even with fairly-moderate deviates from the GT $K$,
DeepDPM's result ($0.66\pm.01$) 
surpasses the leading parametric method. 
Moreover, \autoref{results_figs:balanced_diff_K} shows that the parametric SCAN is sensitive to class imbalance; \eg, in~\autoref{imbalanced}, SCAN performs best when $K = 30$  
suggesting it is due to ignoring  many small classes. In contrast, DeepDPM (scoring $0.60\pm.01$) 
is fairly robust to these changes and is comparable in results to  SCAN when the latter was given the GT $K$. 
In addition, we also show in~\autoref{results:non_param} the performance of other nonparametric methods (3 runs on the same features as ours: MoCo+AE). We include DeepDPM's results with alternation (between clustering and feature learning) and without (\ie, holding the features frozen and training DeepDPM only once). 
\autoref{results:param_non_param} compares the $K$ values found by the nonparametric methods. 
 DeepDPM inferred a $K$ value close to the GT in both the balanced and imbalanced cases. In the imbalanced  case,  moVB scored a \emph{slightly} better $K$ but its results (see~\autoref{results:non_param}) were worse. 
For the parametric methods, \autoref{results:param_non_param} also shows the $K$ value of the best silhouette score. The \emph{unsupervised} silhouette metric is commonly used for model selection (NMI/ACC/ARI are supervised, hence inapplicable for model selection).
As~\autoref{results:param_non_param} shows, DeepDPM yielded a more accurate $K$ than that approach.

\textbf{Running Times.} Our running time is comparable with a \emph{single} run of SCAN (the SOTA deep parametric method); 
\eg, on ImageNet-50, SCAN (with 2 NVIDIA 3090 GPUs) trains for $\sim$8 [hr] while ours
(with 1 weaker NVIDIA 2080 GPU) takes $\sim$11 [hr]. However, \textbf{training SCAN  multiple times with a different $K$ each time (as needed for model selection) took more than 3 days.}  Thus,
DeepDPM's value and positive environmental impact are clear.
\subsection{Ablation Study and Robustness to the Initial $K$}\label{Sec:Ablation}
\autoref{results:DPM_ablations} quantifies the performance gains due to the different parts of DeepDPM through an ablation study done on Fashion-MNIST (in the setting described earlier). It shows the effect of disabling splits, merges and both; \eg, merges help even when initializing with $K=3$. 
In fact, the large moves made by splits/merges help even when $K_\mathrm{init}=10$. 
Also, replacing the subclustering nets with $K$-means (using $K=2$) results in deterioration. 
Likewise, either turning off the priors when computing the cluster parameters, or using an isotropic loss instead of $\Lcal_{\mathrm{cl}}$, hurts performance and (while not shown here) often destabilizes the optimization.  
Finally, \autoref{results:fig_init_k} demonstrates, on three different datasets, DeepDPM's robustness to the initial $K$. 
\begin{figure}[t]
\centering
    \includegraphics[trim={0.3cm 1.15cm 1cm 0.35cm},clip, width=6.5cm]{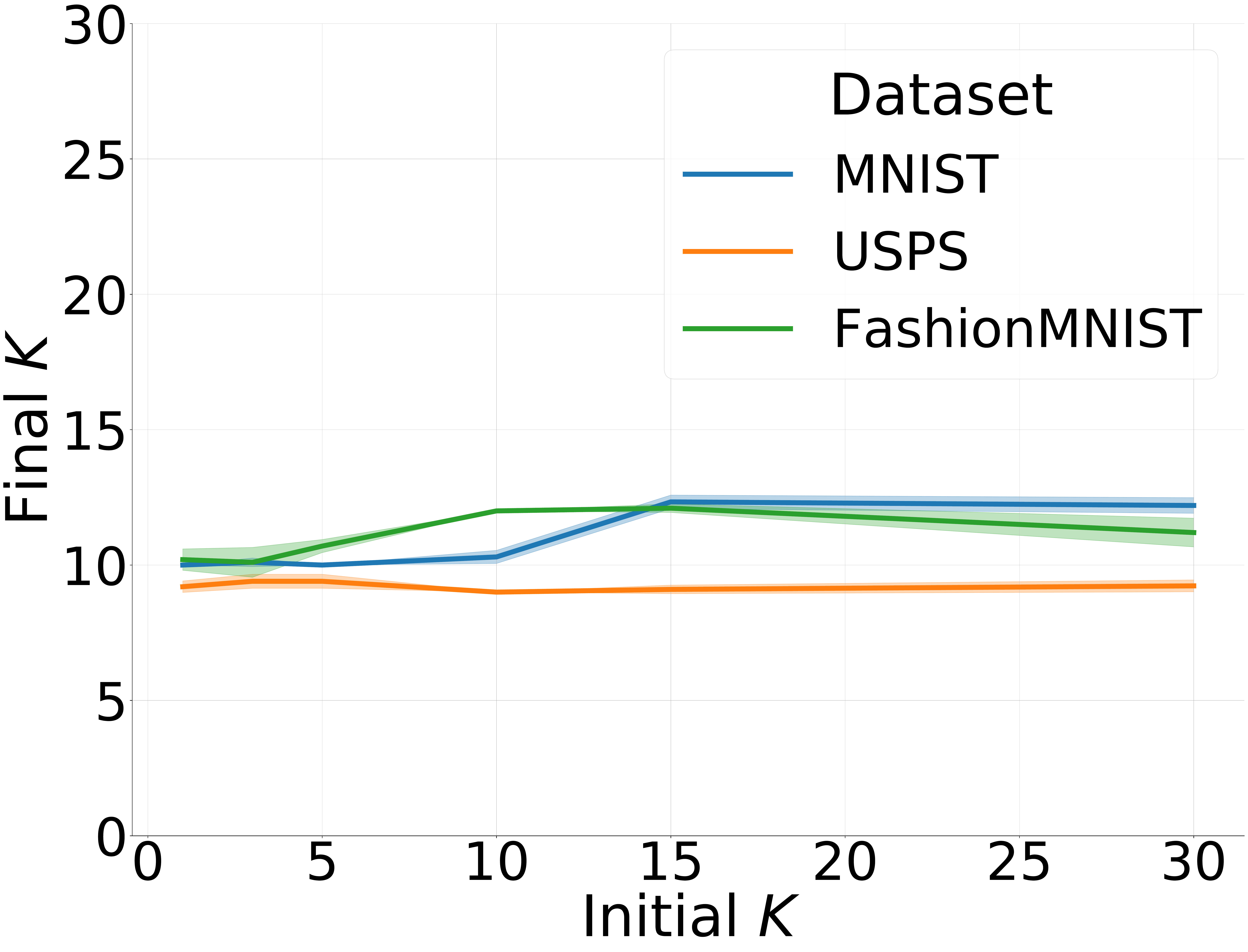}
\caption{Robustness to the initial $K$. GT $K=10$ in all datasets.}
\label{results:fig_init_k}
\end{figure}

\section{Conclusion}\label{Sec:Conclusion}
\textbf{Limitations.}
As with most clustering methods, if DeepDPM's input features are poor it would struggle to recover.
 Also, if $K$ is known and the dataset is balanced, parametric methods (\eg, SCAN) 
may be a slightly better choice. 

\textbf{Future work.}
An interesting direction 
is adapting DeepDPM to streaming data (\eg, similarly to how~\cite{Dinari:AISTATS:2022:Streaming}  handled streaming DPM inference) or hierarchical settings~\cite{Teh:JASA:2006:HDP,Chang:NIPS:2014:ParallelSamplerHDP,Dinari:UAI:2020:vHDP}. 
Moreover, our results may improve given a more sophisticated framework 
for building split proposals (\eg, see~\cite{Winter:AISTATS:2022:splitnet}).

\textbf{Broader impact.} We hope our work will inspire the deep-clustering community to adopt the nonparametric approach as well as raise awareness to issues with the parametric one. 
Nonparametrics also has an environmental positive impact: obviating the need to repeatedly train deep parametric methods for model selection  drastically reduces resource usage. 

\textbf{Summary.}
We presented a deep nonparametric clustering method, a dynamic architecture that adapts to the varying $K$ values, and a novel loss based on new amortized inference in mixture models. Our method outperforms deep and non-deep nonparametric methods and achieves SOTA results. We demonstrated the issues with parametric clustering, especially the sensitivity to the assumed $K$, and the added value the nonparametric approach brings to deep clustering. We showed the robustness of our method to both class imbalance and the initial $K$. Finally, we demonstrated the scalability of DeepDPM by being the first method of its kind to report results on ImageNet. Our code is publicly available.

\clearpage
{\small
\bibliographystyle{./ieee_fullname}
\bibliography{./refs}
}

\end{document}


\title{
DeepDPM: Deep Clustering With an Unknown Number of Clusters \\ ------------ \\  Supplemental Material}

\author{Meitar Ronen \hspace{10em} Shahaf E. Finder \hspace{10em} Oren Freifeld\\
The Department of Computer Science, Ben-Gurion University of the Negev\\
{\tt\small meitarr@post.bgu.ac.il \hspace{8em} \tt\small finders@post.bgu.ac.il \hspace{8em} \tt\small orenfr@cs.bgu.ac.il}
}

\maketitle

This document contains  additional results, implementation details,  and explanations that were omitted from the paper due to page limits. 
\tableofcontents

\section{Additional Results}
\subsection{Additional Visual Examples of Images Clustered Together}
Here, in~\Crefrange{Fig:imgs_clustered_together_sup_1}{Fig:imgs_clustered_together_sup_9}, we provide additional examples of images from the validation set of the ImageNet dataset that were clustered together using DeepDPM. These figures show how DeepDPM grouped images with similar semantic properties.
\twocolumn
\newcommand{\myscale}{0.20}
\begin{figure}
\centering
{\includegraphics[scale=\myscale]{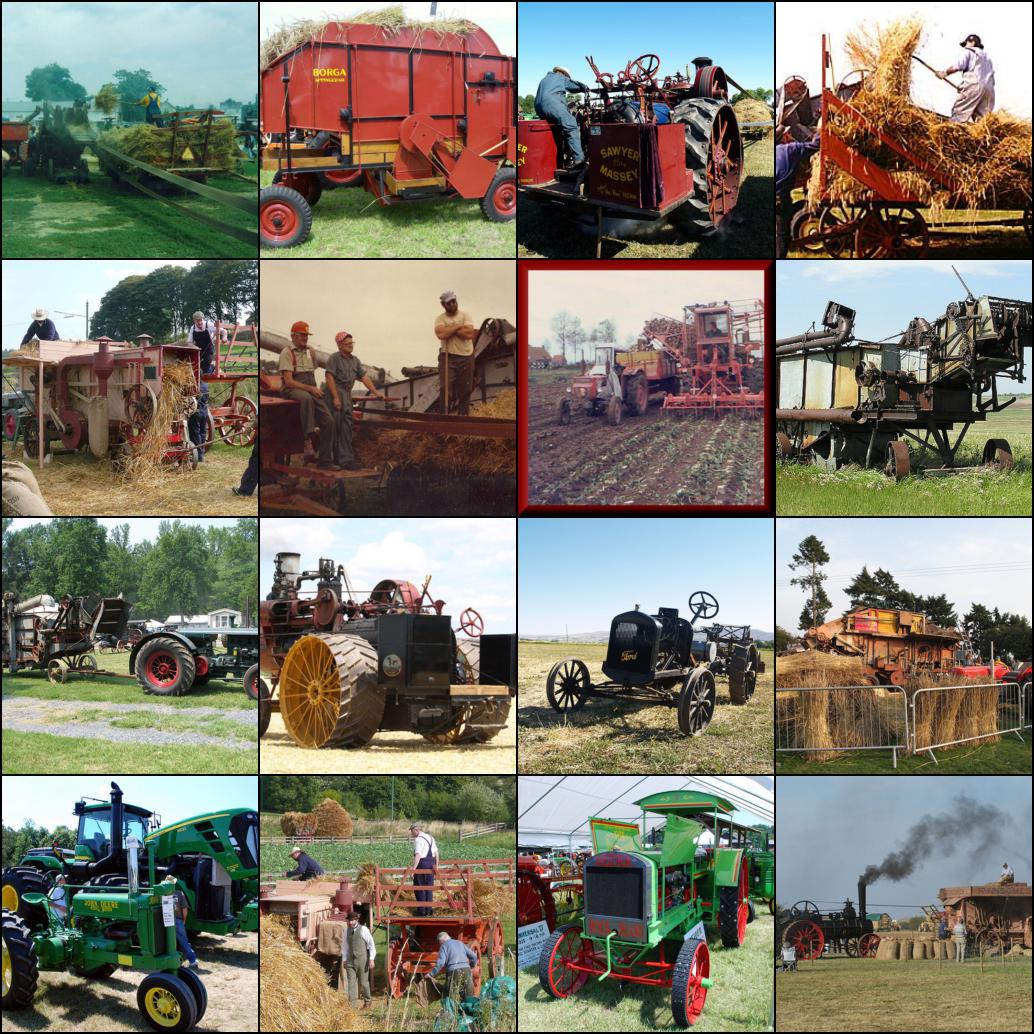}}
\caption{Examples from cluster \#1}
\label{Fig:imgs_clustered_together_sup_1}
\end{figure}

\begin{figure}
\centering
{\includegraphics[scale=\myscale]{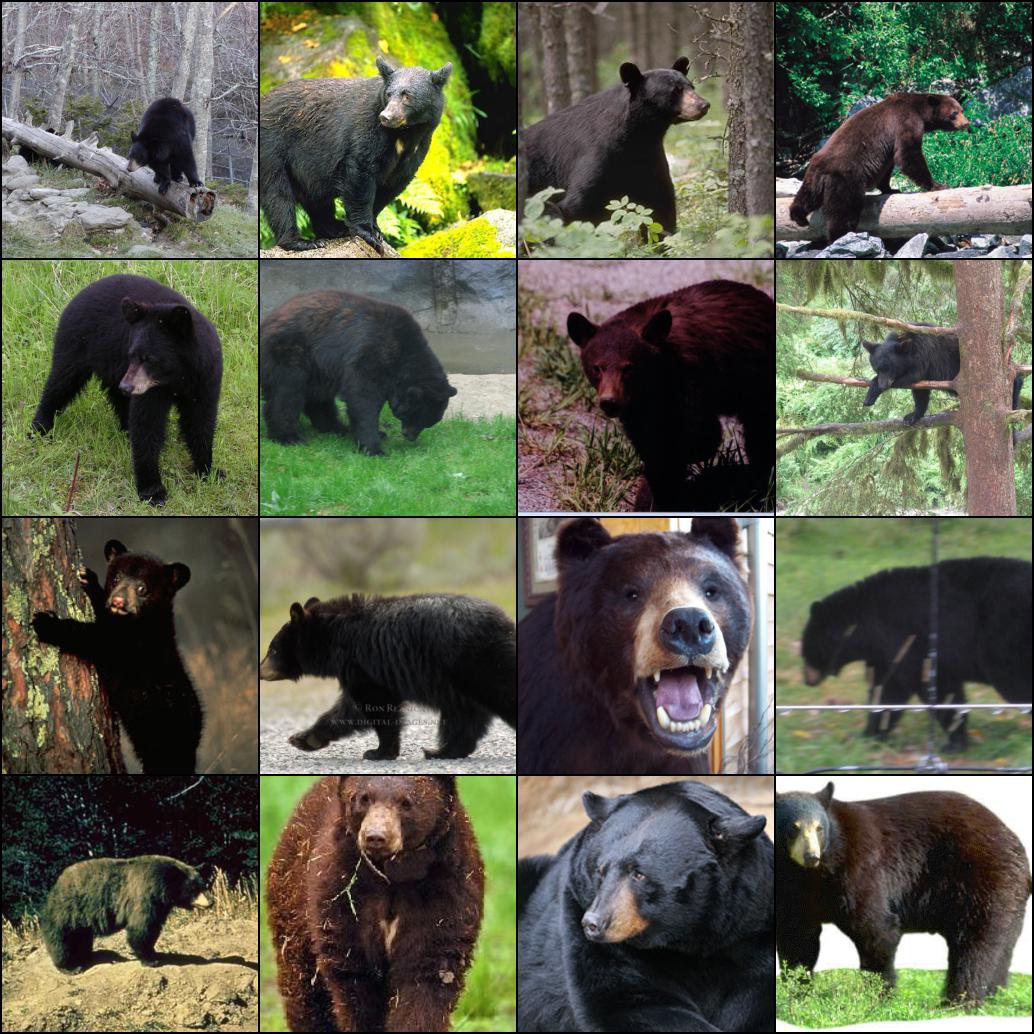}}
\caption{Examples from cluster \#2}
\label{Fig:imgs_clustered_together_sup_2}
\end{figure}

\begin{figure}
\centering
{\includegraphics[scale=\myscale]{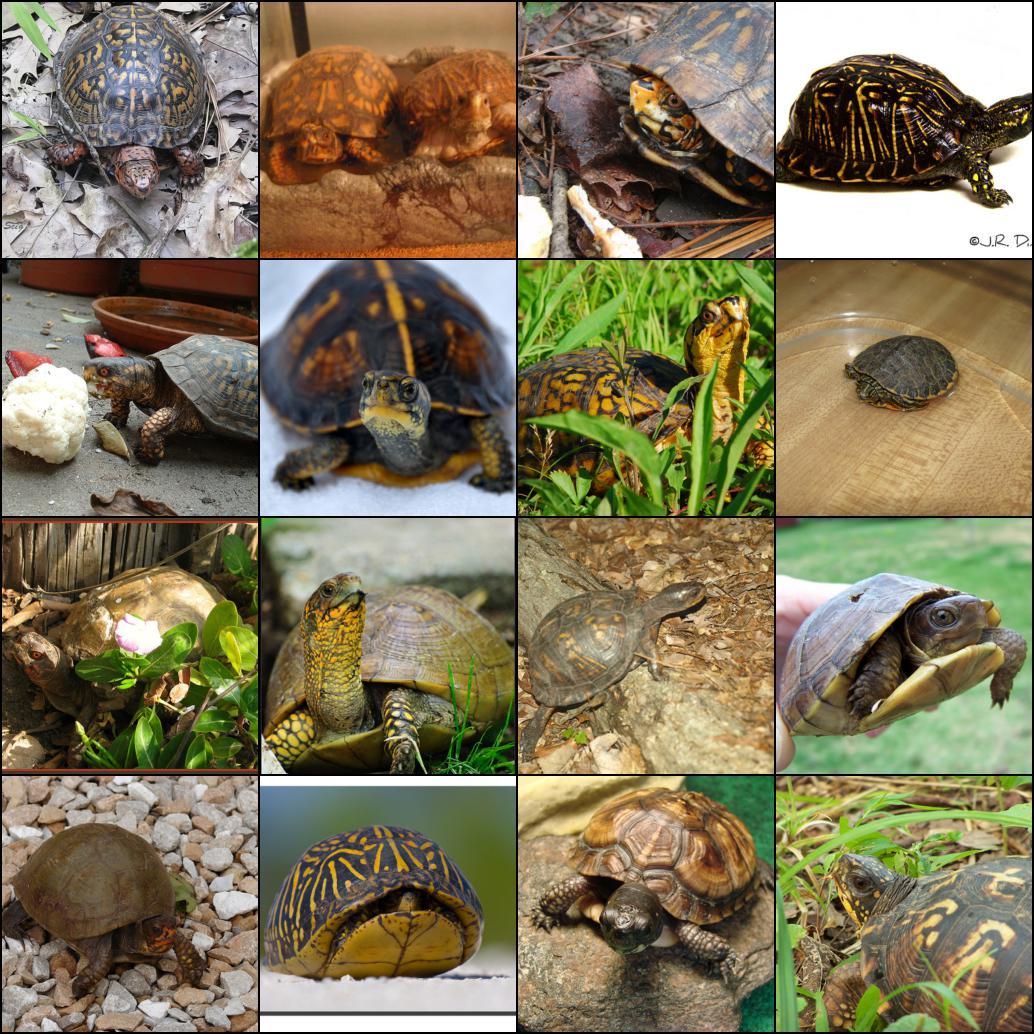}}
\caption{Examples from cluster \#3}
\label{Fig:imgs_clustered_together_sup_3}
\end{figure}

\begin{figure}
\centering
{\includegraphics[scale=\myscale]{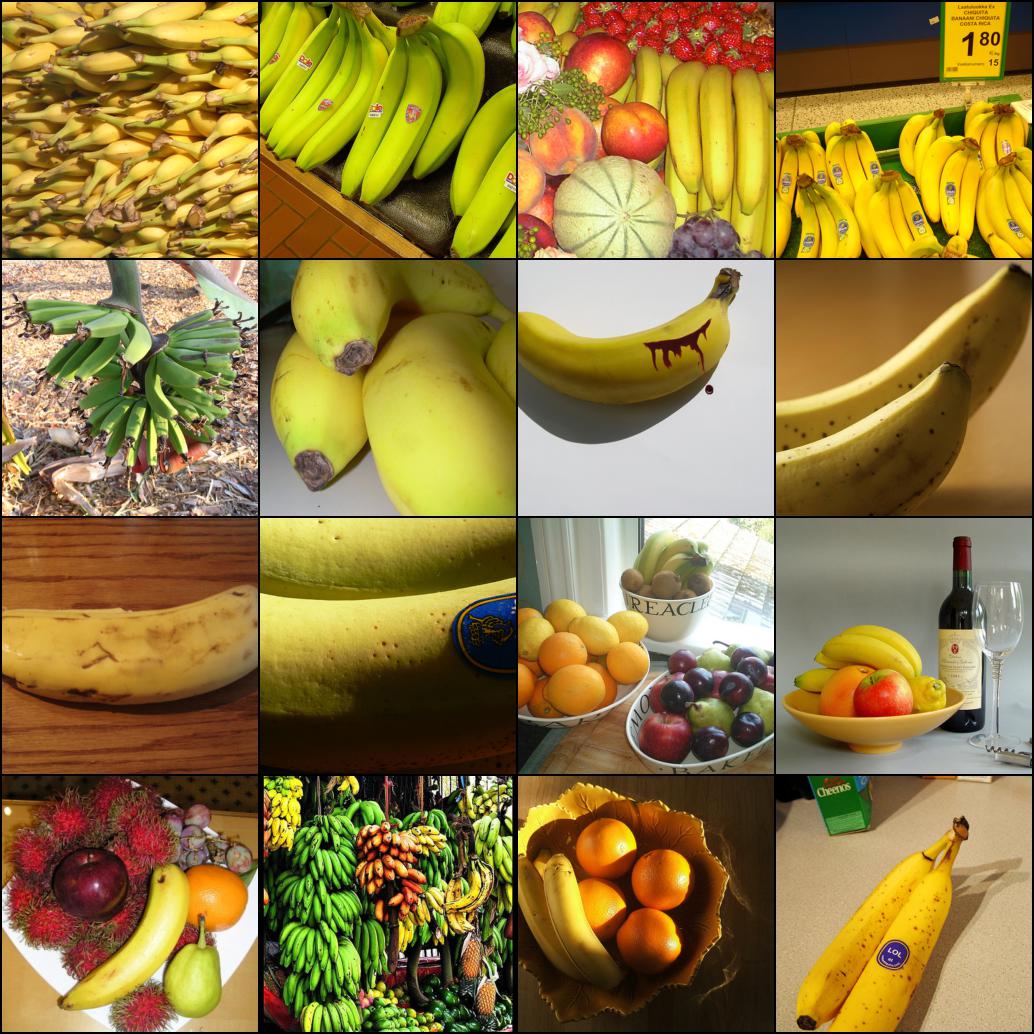}}
\caption{Examples from cluster \#4}
\label{Fig:imgs_clustered_together_sup_4}
\end{figure}

\begin{figure}
\centering
{\includegraphics[scale=\myscale]{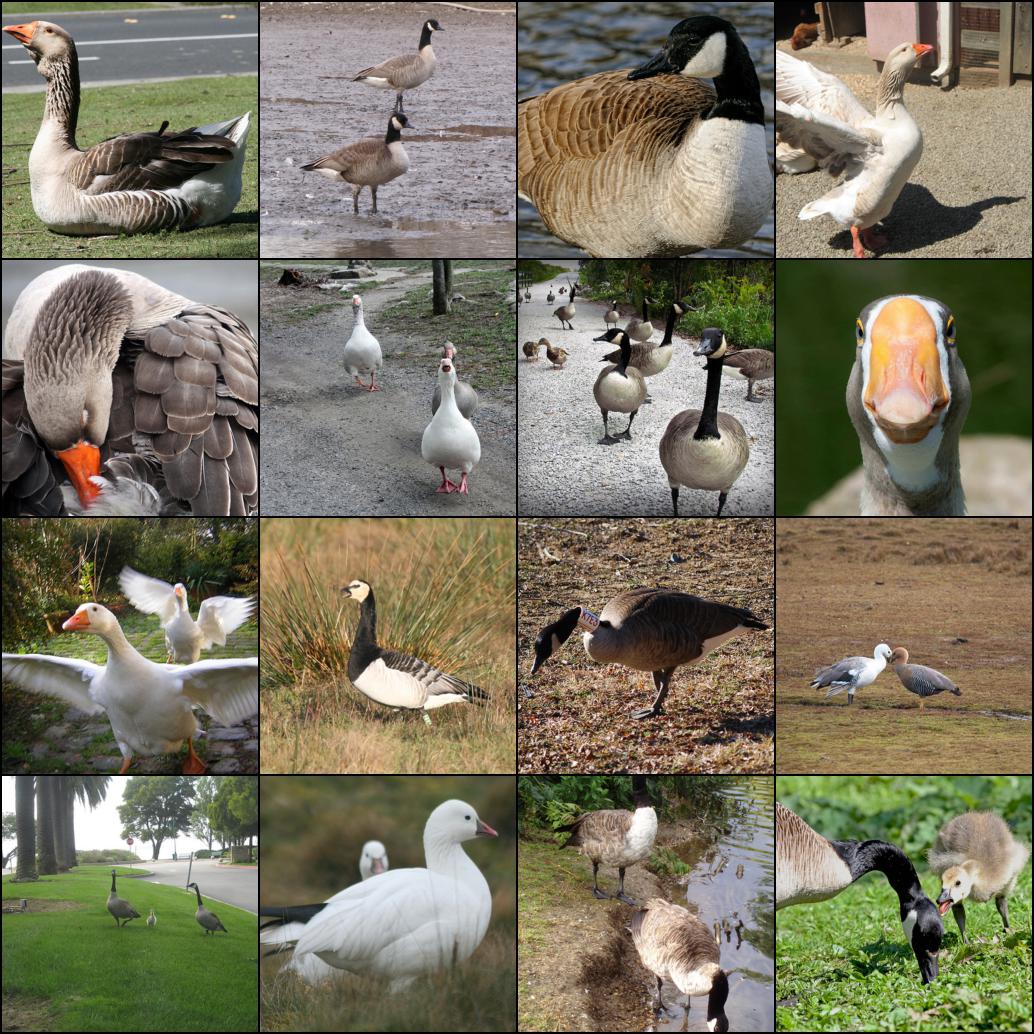}}
\caption{Examples from cluster \#5}
\label{Fig:imgs_clustered_together_sup_6}
\end{figure}

\begin{figure}
\centering
{\includegraphics[scale=\myscale]{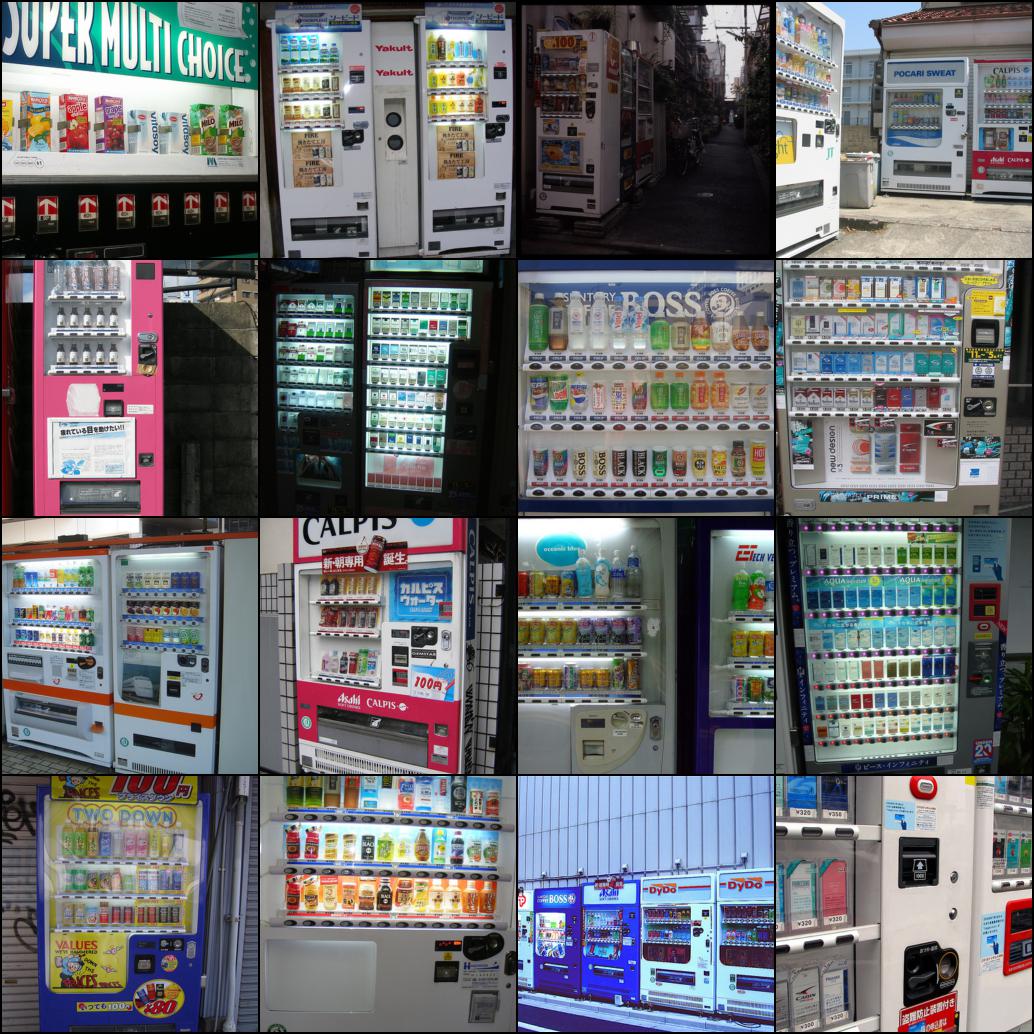}}
\caption{Examples from cluster \#6}
\label{Fig:imgs_clustered_together_sup_7}
\end{figure}

\begin{figure}
\centering
{\includegraphics[scale=\myscale]{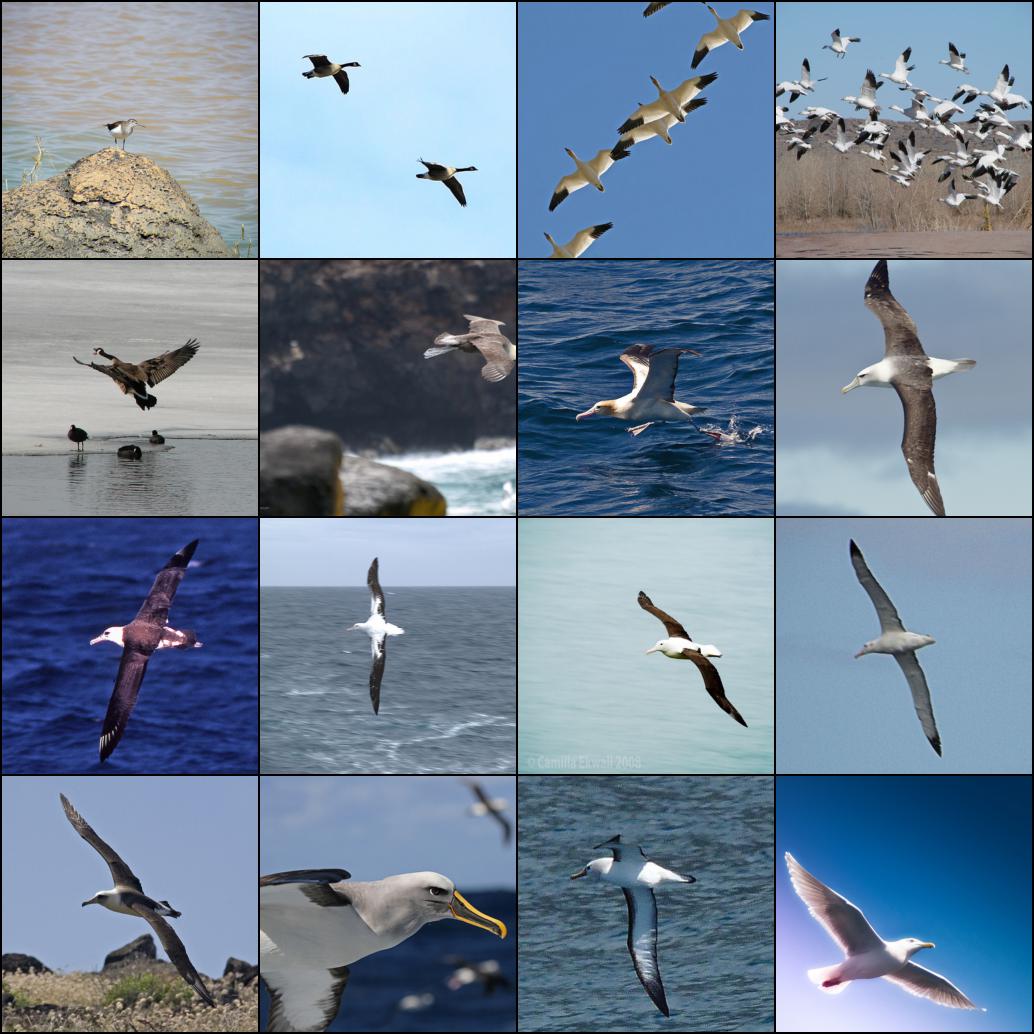}}
\caption{Examples from cluster \#7}
\label{Fig:imgs_clustered_together_sup_8}
\end{figure}

\begin{figure}
\centering
{\includegraphics[scale=\myscale]{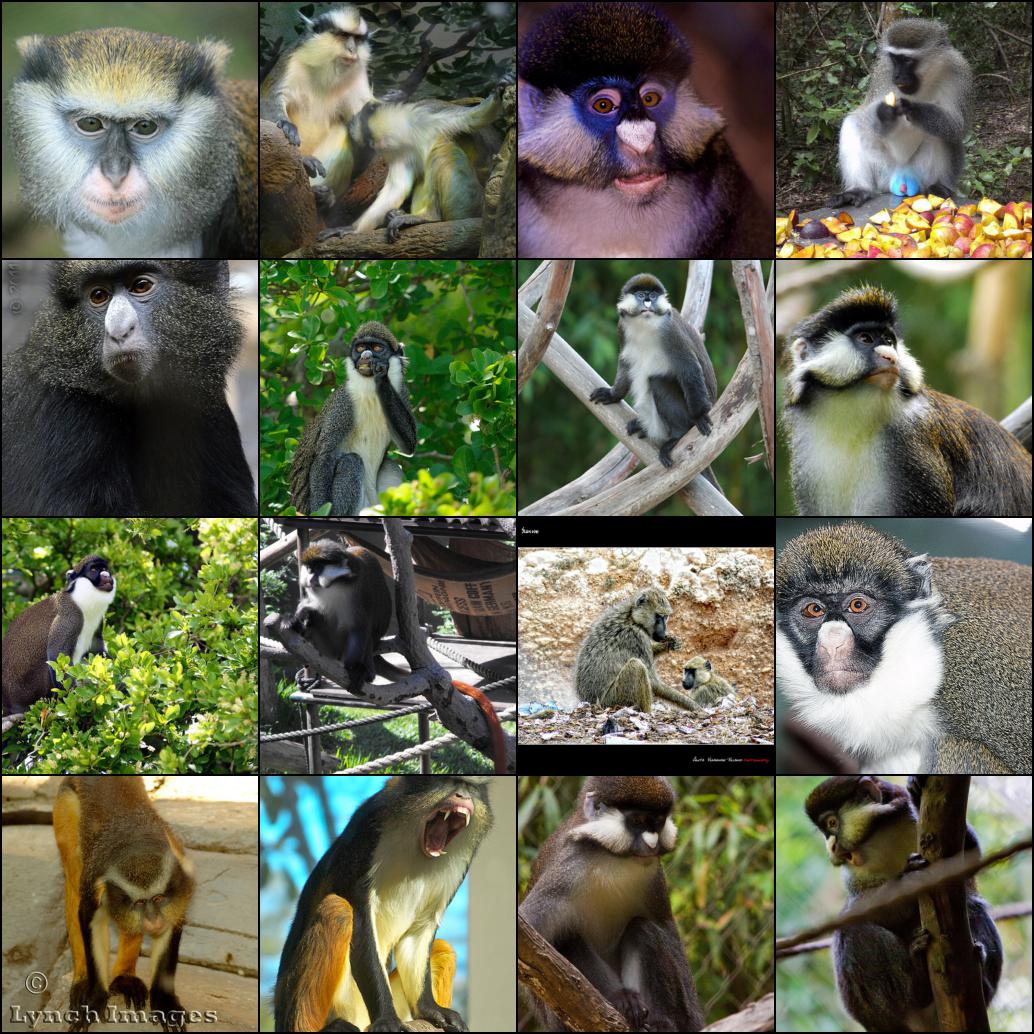}}
\caption{Examples from cluster \#8}
\label{Fig:imgs_clustered_together_sup_9}
\end{figure}

\clearpage
\onecolumn

\subsection{Comparison with Classical Clustering Methods: the Inferred $K$ Value in the Imbalanced Case}\label{results:nonDeep} 
In the paper, we showed how DeepDPM outperforms classical (\ie non-deep) clustering methods, including $K$-means, GMM, DBSCAN~\cite{BNP:Ester:1996:DBSCAN}, moVB~\cite{BNP:Hughes:2013:MoVB} and a State-Of-The-Art (SOTA) DPM sampler~\cite{Dinari:CCGRID:2019:distributed}. We also showed how, in the balanced case, DeepDPM inferred a more accurate estimate of $K$. Here, to complete the picture, we  provide the results for inferring $K$ in the imbalanced setting. In this case too, DeepDPM inferred the most accurate $K$ value; see~\autoref{results:umap_embedded_k_imbalanced}.
\begin{table}[t]
    \centering
\begin{tabular}{@{}p{2.58cm} >{\centering\arraybackslash}p{0.16\linewidth} >{\centering\arraybackslash}p{0.144\linewidth} >{\centering\arraybackslash}p{0.28\linewidth} @{}}
\toprule
Method & \multicolumn{3}{c}{Inferred $K$}\\ 
\midrule
& MNIST$^{imb}$ & USPS$^{imb}$ & Fashion-MNIST$^{imb}$\\
\midrule
DBSCAN & 9.0$\pm$0.00 & 6.0$\pm$0.00 & 4.0$\pm$0.00\\
DPM Sampler & 11.9$\pm$0.32 & 7.3$\pm$0.48 & 11.6$\pm$0.97\\
moVB & 13.6$\pm$0.80 & 11.2$\pm$1.33 & 17.8$\pm$2.27\\
DeepDPM (Ours) & \textbf{10.3$\pm$0.44} & \textbf{9.1$\pm$0.22} & \textbf{9.4$\pm$0.52}\\
\bottomrule
    \end{tabular}
\caption[]{Comparing the mean inferred value ($\pm$std.~dev.) $K$ value, across 10 runs, among
the competing nonparametric methods. GT $K=10$. The symbol $^{imb}$ marks imbalanced datasets.}
\label{results:umap_embedded_k_imbalanced}
\end{table}

\subsection{When Parametric Methods Break: the ARI and NMI Metrics}
\begin{figure}[b!]
\centering
	\subcaptionbox{ImageNet50: The original balanced dataset \label{ARI_balanced}}
		{\includegraphics[scale=.36]{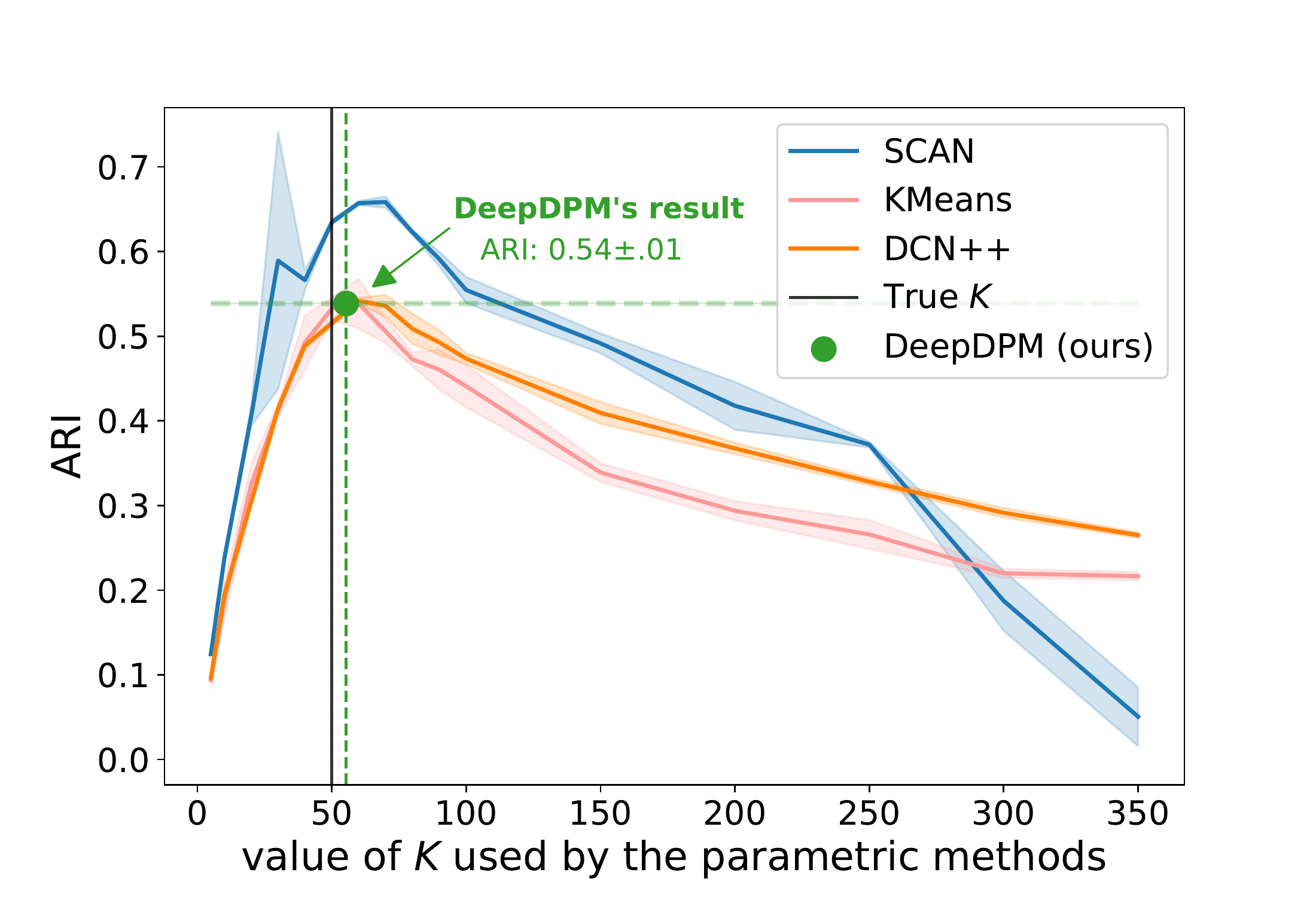}\vspace{-.05cm}}
	\subcaptionbox{ImageNet50: An imbalanced dataset \label{ARI_imbalanced}}
		{\includegraphics[scale=.36]{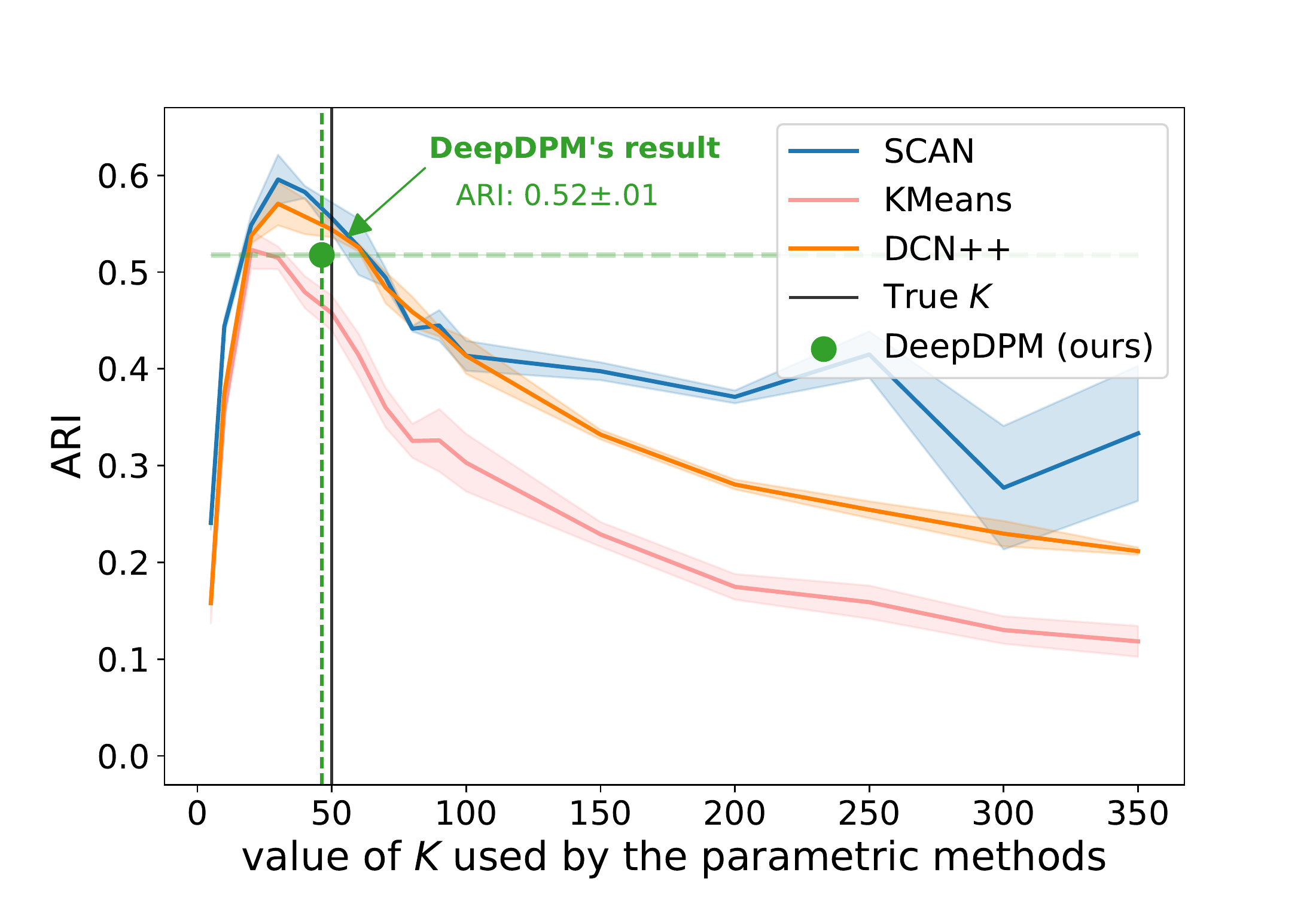}\vspace{-.05cm}}
\caption{Mean ARI of 3 runs ($\pm$ std.~dev.) on 50 classes of ImageNet.
}
\label{results_figs:balanced_diff_K_ARI}
\end{figure}

\begin{figure}[t!]
\centering
	\subcaptionbox{ImageNet50: The original balanced dataset \label{NMI_balanced}}
		{\includegraphics[scale=.36]{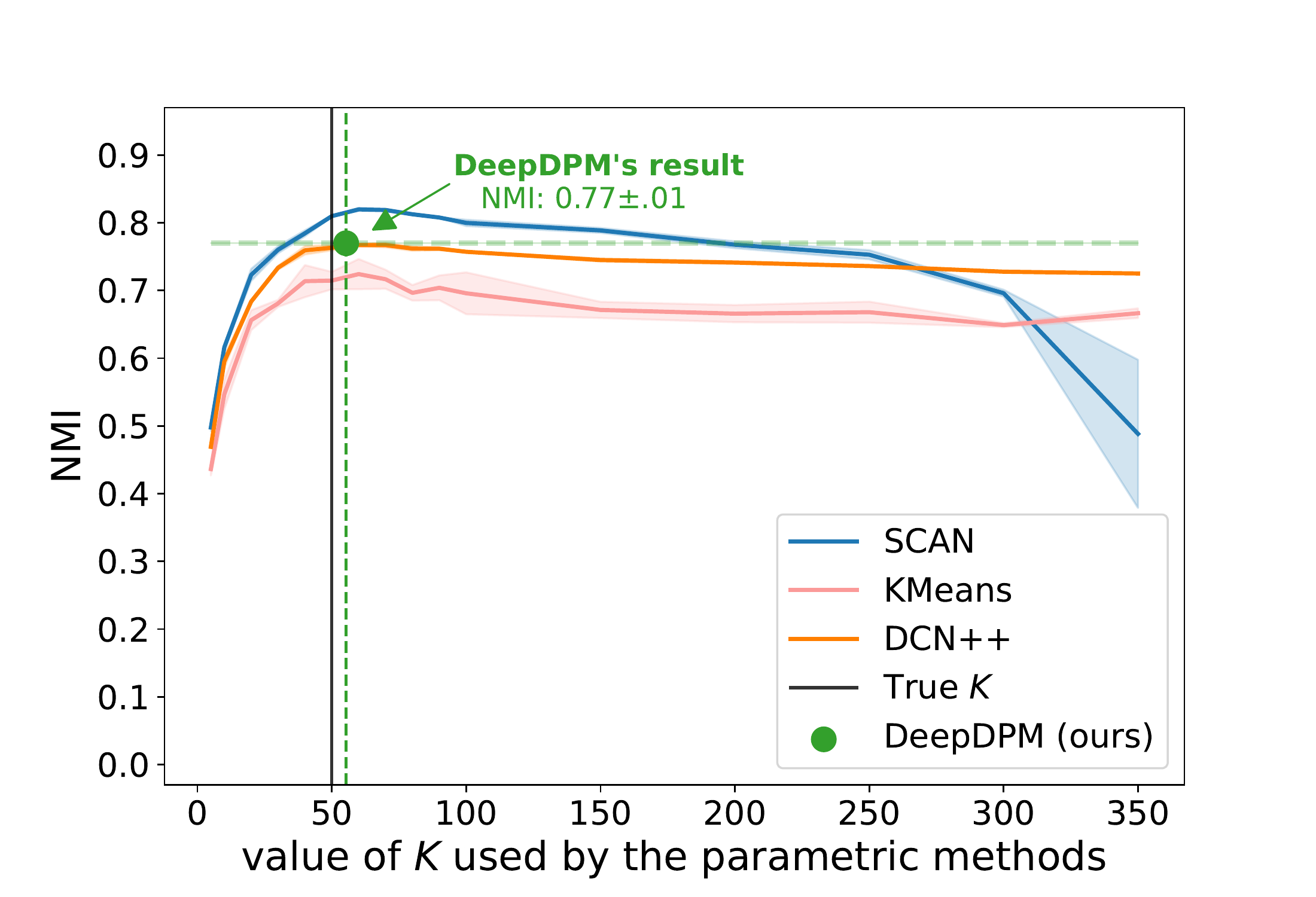}\vspace{-.05cm}}
	\subcaptionbox{ImageNet50: An imbalanced dataset \label{NMI_imbalanced}}
		{\includegraphics[scale=.36]{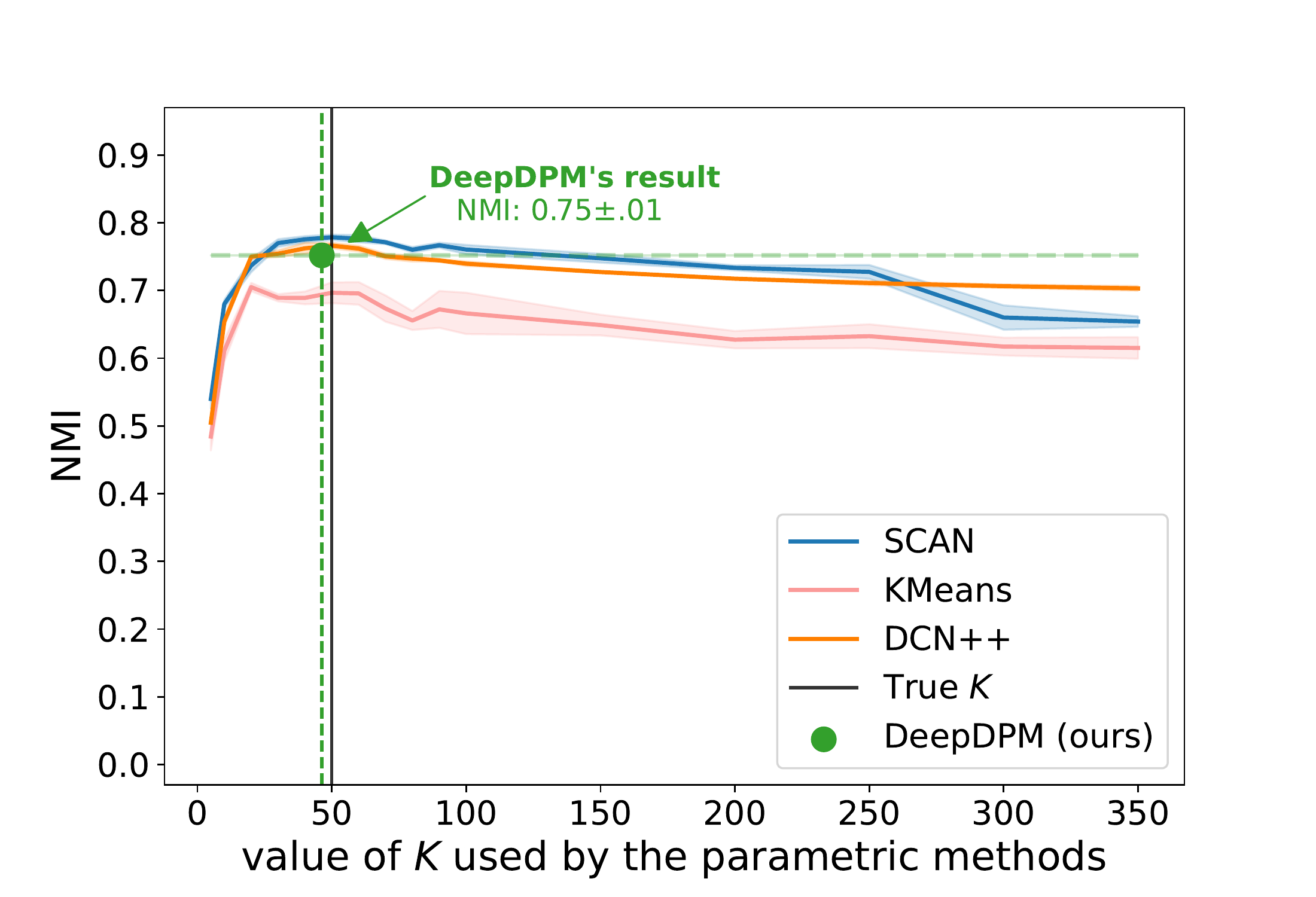}\vspace{-.05cm}}
\caption{Mean NMI of 3 runs ($\pm$ std.~dev.) on 50 classes of ImageNet.
Note that the NMI metric is not sufficiently sensitive to over clustering; \eg, in the balanced case,
SCAN's NMI peaks around $K=70$ while the true $K$ is 50. 
}
\label{results_figs:balanced_diff_K_NMI}
\end{figure}

In the paper, we investigated how parametric methods operate with an unknown $K$, on both balanced and imbalanced datasets. The parametric methods we compared with were $K$-means, DCN++, an improved variant of DCN~\cite{Yang:ICML:2017:towards}, and SCAN~\cite{Gansbeke:ECCV:2020:SCAN}. As the reader may recall, we ran the parametric methods (which require specifying $K$) with different $K$ values  ranging between 5 and 350 (the exact values we used were: 5, 10, 20, 30, 40, 50, 60, 70, 80, 90, 100, 150, 200, 250, 300, and 350). As DeepDPM is a nonparametric method, instead of using a known fixed value of $K$, it infers it. Specifically, on the ImageNet-50 dataset, DeepDPM inferred $K=55.3\pm1.53$ 
(the mean and std.~dev.~across 3 different runs) in the balanced case and $46.3\pm2.52$ in the imbalanced one. In both cases, our results were fairly close to the GT value ($K=50$). 

Here we provide the complementary metrics which were not shown in the paper due to page limits: the ARI (see~\autoref{results_figs:balanced_diff_K_ARI}) and NMI (see~\autoref{results_figs:balanced_diff_K_NMI}) metrics.

While the ARI (similarly to clustering accuracy) penalizes over- and under-clustering similarly, the NMI metric is not as sensitive to over-clustering as it is to under-clustering. For more details, see~\autoref{NMI_explained}. Thus, the NMI score of parametric methods remains relatively-stable when $K \in [50,250]$ for both the balanced case and the imbalanced one. Moreover, as can be seen in~\autoref{NMI_balanced}, for the parametric SOTA method, SCAN, the NMI misleadingly peaks at $K=70$
(recall the true $K$ is 50). 
Despite not having access to the additional information used (and required) by the parametric methods -- that is, the value of $K$ -- and despite the fact that NMI is relatively insensitive to over-clustering, DeepDPM still reaches comparable performance to the parametric methods in both the ARI and NMI metrics, especially in the imbalanced case. 

\subsection{Ablation study: NMI, ARI and the Final Value of $K$}
We provide in~\autoref{results:DPM_ablations_sup} the ARI, NMI and final $K$ values for the ablation study described in the paper. The ACC values already appear in the paper.
\newcolumntype{C}[1]{>{\centering\arraybackslash}p{#1}}

\begin{table*}[h]
\centering
    \begin{tabular}{@{}m{3.755cm} C{1cm}C{1cm}C{1cm} C{1cm}C{1cm}C{1cm} C{1.5cm}C{1.5cm}C{1.5cm} @{}}
    \toprule
 & \multicolumn{3}{c}{NMI} & \multicolumn{3}{c}{ARI} & \multicolumn{3}{c}{final $K$}\\
    \midrule
    & $K_\mathrm{init}$=3 & $K_\mathrm{init}$=10 &$K_\mathrm{init}$=30
    & $K_\mathrm{init}$=3 & $K_\mathrm{init}$=10 &$K_\mathrm{init}$=30
    & $K_\mathrm{init}$=3 & $K_\mathrm{init}$=10 &$K_\mathrm{init}$=30\\
    \midrule
    \hline
    No splits/merges & .53$\pm$.00 & .67$\pm$.01 & .64$\pm$.00& .22$\pm$.00 & .49$\pm$.02 & .43$\pm$.01 & 3 & 10 & 30\\
    No splits & .53$\pm$.00 & .67$\pm$.01 & .63$\pm$.00& .22$\pm$.00 & .49$\pm$.02 & .42$\pm$.02 & 3$\pm$.00 & \textbf{10$\pm$.00} & 23$\pm$.00\\
    No merges & .61$\pm$.00 & .66$\pm$.00 & .64$\pm$.01& .38$\pm$.00 & .48$\pm$.01 & .44$\pm$.01 & 5$\pm$.00 & \textbf{10$\pm$.00} & 21.3$\pm$1.53\\
    2-means instead of $f_{\mathrm{sub}}$
    & \textbf{.68$\pm$.00} & \textbf{.68$\pm$.01} & .67$\pm$.00& \textbf{.50$\pm$.00} & \textbf{.51$\pm$.02} & .48$\pm$.01 & 11$\pm$.00 & 12$\pm$.00 & 14.67$\pm$.58\\
    No priors in the $M$ step
    & .65$\pm$.00 & .66$\pm$.01 & .66$\pm$.00& .48$\pm$.01 & .48$\pm$.02 & .50$\pm$.00 & 12$\pm$1.73 & 14$\pm$1.41 & \textbf{13.67$\pm$.58}\\
    Isotropic loss instead of $\Lcal_{\mathrm{cl}}$ & .67$\pm$.01 & .67$\pm$.00 & .67$\pm$.00 & \textbf{.50$\pm$.01} & .49$\pm$.00 & .49$\pm$.00 & \textbf{10$\pm$0.82} & 9$\pm$.00 & 9.25$\pm$.50\\
	DeepDPM (full method) & \textbf{.68$\pm$.00} & .67$\pm$.01 & \textbf{.68$\pm$.01}& \textbf{.50$\pm$.00} & \textbf{.51$\pm$.01} & \textbf{.52$\pm$.01} & 10.67$\pm$.58 & 11.67$\pm$1.15 & 14$\pm$.00\\
    \bottomrule
    \end{tabular}%
\caption[Ablations]{DeepDPM's performance under different ablations.}
\label{results:DPM_ablations_sup}
\end{table*}

\section{Metrics and Datasets Used in the Evaluation}
\subsection{Evaluation Metrics}
In our evaluations we used three common supervised clustering metrics: clustering accuracy (ACC), Normalized Mutual Information (NMI) and
Adjusted Rand Index (ARI). 
The ACC and NMI range between 0 and 1, and the ARI ranges between -1 and 1. For all metrics the higher the better and all of them can accommodate for different numbers of classes between the prediction and the ground truth.
We also used the silhouette score, which is an unsupervised metric, in order to find (in an unsupervised way) the best $K$ for the parametric methods. The silhouette score ranges between -1 and 1 (the higher the better).  

\subsubsection{ACC}
ACC is defined by:
\begin{align}
	\mathrm{ACC} = \max_m\left(\frac{\sum_{i=1}^N\indicator(y_i = m({z_i}))}{N}\right)
\end{align}
 where $N$ is the number of data points, $y_i$ is the Ground-Truth (GT) class label of data point $i$, $z_i$ is the predicted cluster assignment according to the clustering algorithm under consideration, 
 $\indicator(\cdot)$ is the indicator function, 
 and $m$ is defined by all possible one-to-one mappings between the predicted class membership and the ground-truth one.
 
 Thus, this metric can be compared to the standard accuracy measure used in the supervised-learning settings, with class mapping, where the mapping used is the best match between the GT classes and the predicted ones. To find the best match, we use the popular Hungarian matching algorithm.

\subsubsection{NMI}\label{NMI_explained}
Let $\bz=(z_i)_{i=1}^N$ and let  $\by=(y_i)_{i=1}^N$. 
 NMI is defined by:
\begin{align}
	\mathrm{NMI}= \frac{2 \times I(\by; \bz)}{H(\by) + H(\bz)}
\end{align}
where $H(.)$ stands for entropy and $I(.;.)$ denotes Mutual Information (MI).
One problem with this metric, however, is that the MI term, which appears in the numerator, does not penalize large cardinalities (\ie, over clustering). The denominator partially fixes this,
but not entirely. Thus, NMI is not sensitive enough to over clustering. See for example~\autoref{results_figs:balanced_diff_K_NMI}. 

\subsubsection{ARI}
The Rand index (RI) quantifies the percentage of ``correct'' decisions for each pair of data points. A decision is correct if two examples either belong to the same GT class and the same cluster assignment (a true positive, TP), or being from two different GT classes and assigned to two different clusters (a true negative, TN). Similarly, clustering errors are false positives (FP) and false negatives (FN). Then RI is computed by:
\begin{align}
	RI = \frac{TP + TN}{TP + TN + FP + FN}.
\end{align}

The ARI measure is the corrected-for-chance version of the Rand index. Given a set $\mathcal{S}$ of $N$ elements, and two groupings or partitions (e.g. $\by$ and $\bz$) of these elements, the overlap between $\by$ and $\bz$ can be summarized in a contingency table $[c_{kl}]$ where each entry $c_{kl}$ denotes the number of objects in common between $y_k$ and $z_k : c_{kl} = |y_k \cap z_k|$. Let $a_k$ be the sum of each row, meaning, $a_k = \sum_{l}c_{kl}$, and $b_k$ the sum of each column, \ie $b_k = \sum_{k}c_{kl}$.

The ARI measure is then calculated by:
\begin{align}
	ARI = \frac{\sum_{kl} {n_{kl} \choose 2} - [\sum_k {a_k \choose 2} \sum_l{b_l \choose 2}] / {n \choose 2} }{\tfrac{1}{2} [\sum_k {a_k \choose 2} + \sum_l{b_l \choose 2}] - [\sum_k {a_k \choose 2} \sum_l{b_l \choose 2}]/ {n \choose 2}}
\end{align}

\subsubsection{Silhouette Score}

So far we have discussed supervised evaluation scores, meaning, ones that require the GT labels. However, in unsupervised cases where the GT is unknown in training, one often needs a metric to evaluate the model; \eg when performing hyperparameter tuning or model selection for parametric methods. In this case, usually a parametric model is run using a range of different values for $K$, and an unsupervised criterion is used to choose the best model (best value for $K$).

One of the most common unsupervised clustering metrics is the silhouette score, which quantifies the clustering quality by measuring the amount of ``cohesion'' within a cluster, and ``separation'' between different clusters. Meaning, the more data points within each cluster are closely-packed and different clusters are well-separated, the higher the silhouette score is. More formally, given data $\mathcal{X} = (\bx_i)_{i=1}^N \in \mathcal{R}^{N \times d}$ and its clustering prediction $\bz$, for data point $\bx_i$ with cluster label $k$ ($z_i= k$), let

\begin{align}
	a(i) = \frac{1}{|N_{k}| -1}\sum_{\bx_j: z_j = k}d(\bx_i, \bx_j)
\end{align}
be the mean distance between $\bx_i$ and all other data points in the same cluster, where $|N_{k}|$ is the number of points hard assigned to cluster $k$, and and $d(i,j)$ is the distance between data points $\bx_i$ and $\bx_j$.

Let
\begin{align}
	b(i) = \min_{k': k' \neq k}\frac{1}{N_{k'}}\sum_{\bx_j: z_j = k'}d(\bx_i, \bx_j)
\end{align}

be the smallest mean distance of datapoint $\bx_i$ to all points in any other cluster, of which $\bx_i$ is not a member.

Now, the silhouette score of $\bx_i$ is defined as:

\begin{align}
s(i) = 
\begin{cases}
      \frac{b(i) - a(i)}{\max(a(i), b(i))}, & \text{if}\ N_{k} > 1 \\
      0, & \text{otherwise}
    \end{cases}
\end{align}

Thus, $s(i) \in [-1, 1]$.
Finally, the total silhouette score is the average of all values for $s(i)$.
Note that the silhouette score does not use the GT labels, and thus it is an unsupervised metric.

\subsection{Datasets for Evaluation}

\textbf{Datasets.} We evaluate our method on text and image datasets in varying scales. The summary statistics on the datasets are available in~\autoref{results:datasets_summary}.
\begin{table*}[ht]
\centering
\begin{tabular}{|c|c|c|c|c|}
\hline
\textbf{Dataset} & \textbf{Train samples} & \textbf{Val samples} & \textbf{Data Dimension} &\textbf{GT $K$}\\
\hline 
MNIST~\cite{Deng:2012:Mnist} & 60,000 & 10,000 & 28 $\times$ 28 & 10\\
\hline 
USPS~\cite{uspsdataset} & 7291 & 2007 & 16 $\times$ 16 & 10\\
\hline
Fashion- MNIST~\cite{Xiao:2017:Fashion-MNIST} & 60,000 & 10,000 & 28 $\times$ 28 & 10\\
\hline
STL10~\cite{Coates:AISTATS:2011:STL-10} & 5,000 & 8,000 & 96 $\times$ 96 $\times$ 3 & 10\\
\hline
Reuters10K~\cite{Lewis:2004:rcv1} & 10000 & - & 28 $\times$ 28 & 4\\
\hline
ImageNet-50 &  64,274 & 2,500 & 224 $\times$ 224 $\times$ 3 & 50\\
\hline
ImageNet~\cite{Deng:CVPR:2009:imagenet} & 1,281,167 & 50,000 & 224 $\times$ 224 $\times$ 3 & 1000\\
\hline
\end{tabular}
\caption[Descriptive properties of the datasets used for evaluation]{Descriptive properties of the datasets used for evaluation.}
\label{results:datasets_summary}
\end{table*}

\textbf{Remarks regarding the ImageNet and ImageNet-50 datasets.}
The creators of ImageNet~\cite{Deng:CVPR:2009:imagenet} do not hold the copyright of all images, and the usage of that dataset is governed by the terms of the ImageNet license \url{https://image-net.org/download}.
ImageNet-50 is a subset of 50 randomly-selected classes of ImageNet, curated by~\cite{Gansbeke:ECCV:2020:SCAN}.

\subsection{Train and Validation Sets} In general, we train and evaluate using the train and validation split respectively for most of the comparisons. However, when comparing with deep nonparametric methods (which we had problems to run their code and had to resort to use their reported results), to allow for a fair comparison, we computed the evaluation metrics on the entire dataset (as this is what their reported numbers referred to), meaning, combining the train and validation sets into one set. 
Note that in this case too, we still trained our model only on the training set. 
Also recall the training (of both our method and the competitors) is unsupervised
and used neither the GT labels nor the GT $K$. 

\subsection{Creating Imbalanced Datasets}

To create imbalanced datasets for the smaller datasets, \ie, MNIST, USPS and Fashion-MNIST, we undersampled some of the classes using random proportions.
Concretely, for MNIST and USPS we sampled 10\%, 5\%, 20\%, and 30\% of the total amount of examples of 
classes 8, 2, 5, and 9, respectively. All the other classes were used in full.
For Fashion-MNIST dataset, classes 0, 3, 5, 7, and 8 were under-sampled with 37\%, 19\%, 41\%, 54\%, 19\% of the total number of examples per class (the classes and percentages were  chosen randomly).\\

\begin{figure}[h!]
\centering
\subcaptionbox{The random histogram we sampled from the uniform distribution over the 50-dimensional probability simplex. This histogram was used for creating the imbalanced version of ImageNet-50.\label{results_figs:imb_imagenet50_alpha_1}
}%
{\includegraphics[scale=.6]{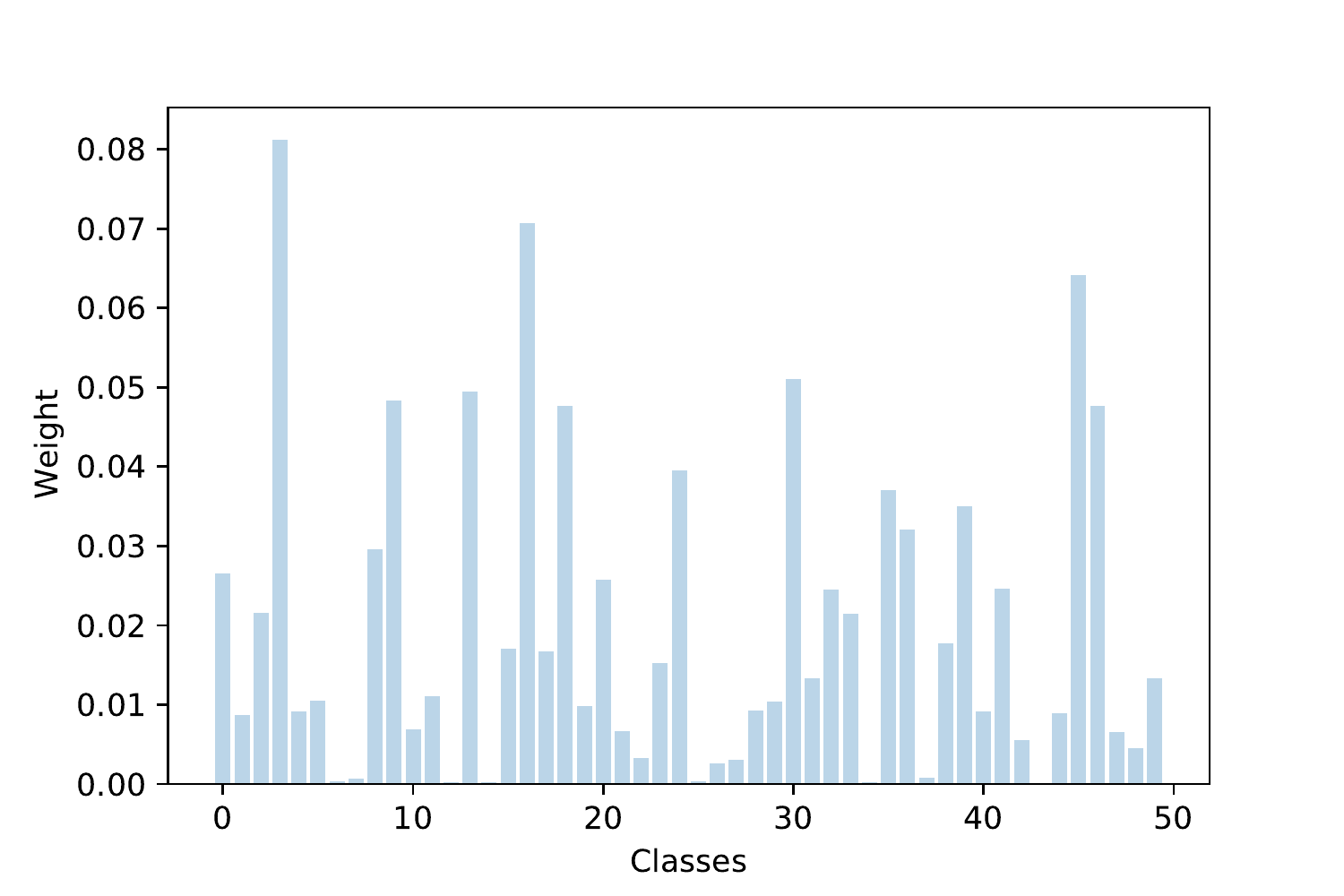}}%
\subcaptionbox{The class histogram of the original ImageNet-50 dataset. \label{results_figs:imagenet50_hist}}
{\includegraphics[scale=.6]{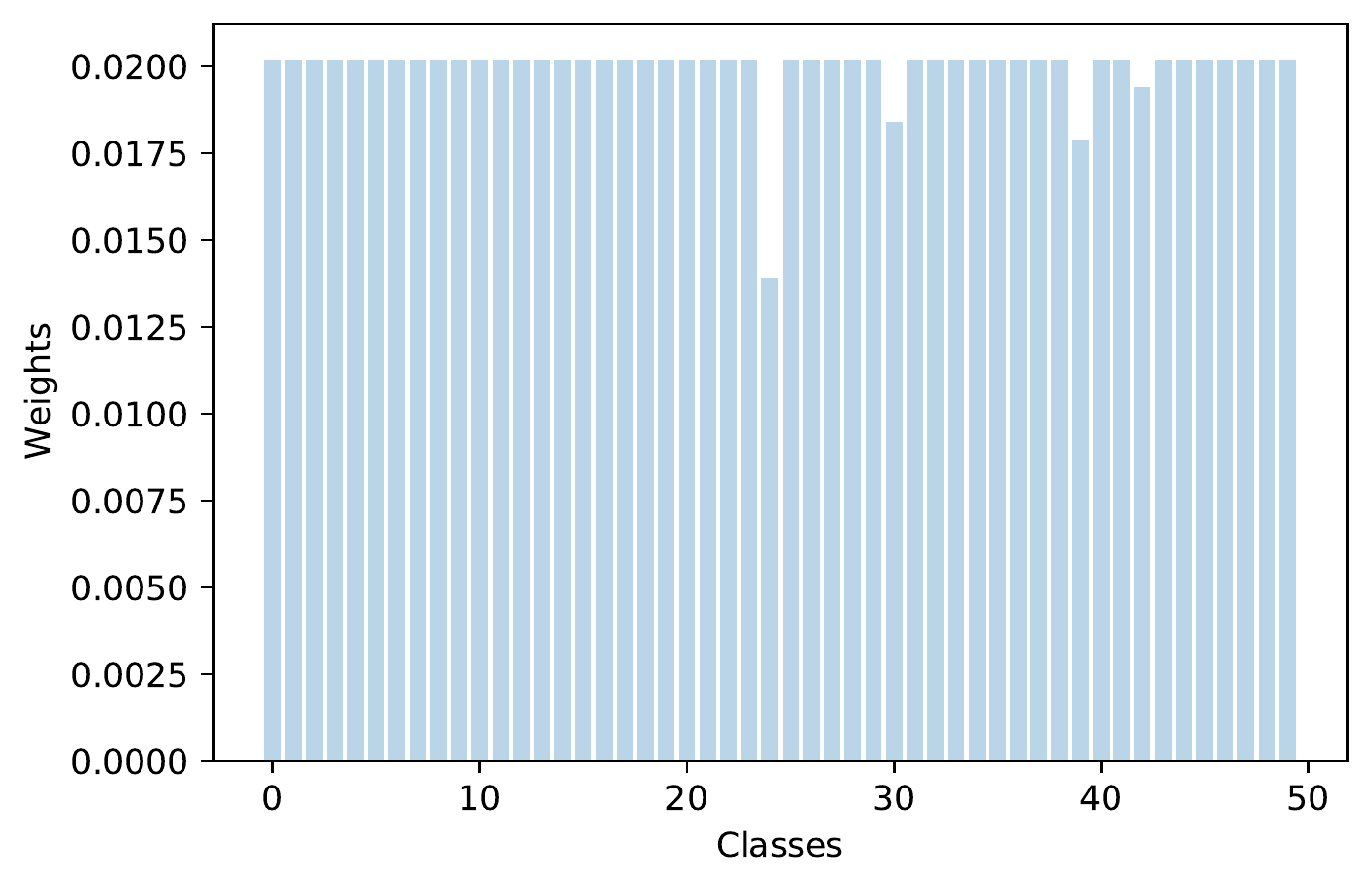}}
\caption{The balanced vs. imbalanced histograms for ImageNet-50.}
\label{Fig:balanced_imbalanced_imagenet-50}
\end{figure}

To create an imbalanced version of ImageNet-50, 
we sampled a 50-bin normalized histogram from a uniform Dirichlet distribution 
(not to be confused with a Dirichlet process) 
over the 50-dimensional probability simplex. This means that any histogram was equally probable (not to be confused with a uniform histogram). The random
nonuniform histogram we sampled is shown in~\autoref{results_figs:imb_imagenet50_alpha_1}.  For comparison, \autoref{results_figs:imagenet50_hist} shows the original class distribution of ImageNet-50 
which is almost perfectly balanced (\ie, almost uniform). 

\section{The NIW Prior, Marginal Data Likelihood, the Weighted Bayesian Estimates, and the Concentration Parameter}
To make our work self-contained, below we provide the details for the key Bayesian calculations that we use. For more details about the known theoretical results in~\autoref{Sec:NIW},~\autoref{Sec:Marginal}, and~\autoref{Sec:DP}, see~\cite{Chang:Thesis:2014:Sampling}.  

\subsection{The Normal Inverse Wishart Distribution}\label{Sec:NIW}
In the Dirichlet Process Gaussian Mixture Model (DPGMM), like in the Bayesian GMM, each component's parameters, $(\btheta_k, \pi_k)$, where $\btheta_k = (\bmu_k, \bSigma_k)$ denote the mean and covariance matrix, and $\pi_k$ is the mixture weight, are assumed to be drawn from a certain prior distribution.
A common choice for a prior for $\btheta_k$ is the Normal-Inverse-Wishart (NIW) distribution. This is because the latter is a conjugate priorto the multivariate normal distribution with an unknown mean and an unknown covariance matrix. The conjugacy property guarantees that the posterior probability will be in the same distribution family as the NIW prior, and provides a closed-form expression for it, which is algebraically convenient for inference.\\

The probability density function (pdf) of the Inverse-Wishart (IW) distribution over $d\times d$ Symmetric and Positive Definite (SPD) matrices, 
evaluated at the $d\times d$ SPD matrix $\bSigma_k$, 
is 
\begin{align}
\Wcal^{-1}(\bSigma_k;\nu,\bPsi)=
\frac{\left|\nu{\bPsi}\right|^{\frac{\nu}{2}}}{2^{\frac{\nu 
d}{2}}\Gamma_d(\frac{\nu}{2})} 
\left|\bSigma_k\right|^{-\frac{\nu+d+1}{2}}e^{-\frac{1}{2}\operatorname{tr}({
\nu\bPsi}\bSigma_k^{-1})}
\end{align}
where $\nu>d-1$, $\bPsi\in \RR^{d\times d}$ is SPD,
and $\Gamma_d$ is the ($d$-dimensional) multivariate gamma function. 
Equivalently, we may write 
\begin{align}
 \bSigma_k\sim \Wcal^{-1}(\nu,\bPsi) \,. 
\end{align}
The positive real number $\nu$ and the SPD matrix $\bPsi$ are called the hyperparameters of the IW distribution. 

Now let $\bmu_k\in \Rd$. The vector $\bmu_k$ and the SPD matrix $\bSigma_k$ are said to be Normal-Inverse-Wishart distributed 
        if their joint pdf is 
        \begin{align}
        &p(\bmu_k,\bSigma_k;\kappa,\bm,\nu,\bPsi)
        =        
       \mathrm{NIW}(\bmu_k,\bSigma_k;\kappa,\bm,\nu,\bPsi)
       \triangleq         
\overbrace{\Ncal(\bmu_k;\bm,\tfrac{1}{\kappa}\bSigma_k)}^{p(\bmu_k|\bSigma_k;\kappa,
\bm)} 
\overbrace{\Wcal^{-1}(\bSigma_k;\nu,\bPsi)}^{p(\bSigma_k;\nu,\bPsi)}
        \end{align}       
where 
$\bm\in \Rd $ and $\kappa>0$ (while $\nu$ and $\bPsi$ are as before)
and $\Ncal(\bmu_k;\bm,\tfrac{1}{\kappa}\bSigma_k)$ is a $d$-dimensional Gaussian pdf,
evaluated at $\bmu_k$, with mean $\bm$ and covariance $\tfrac{1}{\kappa}\bSigma_k$.

Equivalently, we may write 
\begin{align}
	(\bmu_k, 	\bSigma_k) \sim \mathrm{NIW}(\bm, \kappa, \bPsi, \nu) \, . 
\end{align}

The elements of the tuple $\lambda\triangleq(\bm,\kappa,\bPsi,\nu)$ are called the hyperparameters of the NIW distribution. Particularly, $\nu$ and $\kappa$ are called the pseudocounts of that distribution.
Loosely speaking, the higher $\nu$ and $\kappa$ are, the more the distribution is peaked
(roughly) around $\bPsi$ and around $\bm$, respectively. 
\\

\textbf{Remark}: do not confuse $k$ (the index of the Gaussian/cluster) with $\kappa$ (``kappa'', a hyperparameter of the NIW distribution). \\

Assuming, for a moment, a hard-assignment setting, let $N_k=|\set{i:z_i=k}|$ and let $\Xcal_k=(\bx_i)_{i:z_i=k}$ denote  $N_k$ \iid draws from $\Ncal(\bmu_k,\bSigma_k)$. 
The key reason why the NIW distribution is widely used~\cite{Gelman:Book:2013:Bayesian} as a prior over the parameters, $(\bmu_k,\bSigma_k)$, is
conjugacy (to the Gaussian likelihood). Namely, the posterior distribution over these parameters is also NIW, 
\begin{align}
 p(\bmu_k,\bSigma_k|\Xcal_k)= \mathrm{NIW}(\bmu_k,\bSigma_k;\kappa^*,\bm_k^*,\nu^*,\bPsi_k^*) \, ,
\end{align}
and its so-called posterior hyperparameters are given in closed form:
        \begin{align}
         \kappa_k^* &= \kappa + N_k  \label{Eqn:NIWposteiorUpdates1}\\
         \qquad \bm_k^* &=    
             \frac{1}{\kappa_k^*}
         \left[\kappa \bm + \sum_{i:z_i=k} \bx_i   \right]
        \label{Eqn:NIWposteiorUpdates2}\\
         \nu_k^* &= \nu + N_k 
        \label{Eqn:NIWposteiorUpdates3}\\
         \bPsi_k^*&=
         \frac{1}{\nu^*}
         \left[
         \nu \bPsi +\kappa \bm\bm^T 
         +\left(\sum_{i:z_i=k}\bx_i\bx_i^T\right)- \kappa_k^*\bm_k^*(\bm_k^*)^T 
         \right] \ . \label{Eqn:NIWposteiorUpdates4}
        \end{align}
Importantly, when $\nu$ and $\kappa$ are much smaller than $N$, then the specific choice of $\bmu_k$ and $\bPsi$ becomes negligible, implying a very weak prior.

\subsection{The Marginal Likelihood Function}\label{Sec:Marginal}
When marginalizing over the parameters of a Gaussian (\ie, its mean and covariance),
one obtains the marginal data likelihood (given the hyperparameters of the NIW prior): 
\begin{align}
    &  f_\bx((\bx_i)_{i=1}^N;\lambda)=
    f_\bx((\bx_i)_{i=1}^N;\bm,\kappa , \bPsi ,\nu) =
    \int p((\bx_i)_{i=1}^N|\bmu_k,\bSigma_k)p(\bmu_k,\bSigma_k ;\lambda) d(\bmu_k,\bSigma_k) 
    \nonumber \\ &=
    \frac{1}{\pi^{\frac{N d}{2}}} \frac{\Gamma_{d}\left(\nu^* / 2\right)}{\Gamma_{d}(\nu / 2)} \frac{|\nu \bPsi|^{\nu / 2}}{\left|\nu^* \bPsi_k^*\right|^{\nu^* / 2}}\left(\frac{\kappa}{\kappa^*}\right)^{d / 2} \label{Eqn:LL_post} \, 
\end{align}    
where $\Gamma_d$ is the $d$-dimensional Gamma function.

\subsection{Weighted MAP Estimates of the Parameters}\label{Sec:WeightedMAP}
We now provide the details of the M step. More concretely, below we explain how we compute the weighted Maximum-a-Posteriori (MAP) estimates of the clusters' and subclusters' parameters,
 where the weighting is done according to the output of our deep nets.

Let $\lambda = (\bm, \kappa, \bPsi, \nu)$ be the NIW hyperparams.
In the unweighted case, the MAP estimates of $\bmu$ and $\bSigma$ are:
\begin{align}
 {\bSigma_k}&= \frac{\nu^* \bPsi_k^*}{\nu^*-d+1} \label{Eqn:MAPmean}\\
 {\bmu_k} &= \bm_k^*  \, . \label{Eqn:MAPcov}
\end{align}
In our case, the MAP estimates are obtained in a similar way, but with the following differences. 
Rather than using hard assignments (as in~\autoref{Sec:NIW}), 
we use soft assignments; \ie, the MAP estimates take all $N$ points into consideration, but with an appropriate weighting of each point. This is nearly identical to the
weighted MAP estimates in the standard computation of the M step in Bayesian EM-GMM, except that here the weighting is done using the $(r_{i,k})$ values (namely, the soft assignments which are our deep net's output). 
That is,  we still use \EQN\eqref{Eqn:MAPmean}
and \EQN\eqref{Eqn:MAPcov}, but instead of the posterior hyperparameters from 
\EQNS\eqref{Eqn:NIWposteiorUpdates1}--\eqref{Eqn:NIWposteiorUpdates4}, 
we use their weighted versions (where the weights are the $(r_{i,k})$ values). 
 \begin{align}
         \kappa_k^* &= \kappa +\sum_{i=1}^N r_{i,k}  \\
         \qquad \bm_k^* &=    
             \frac{1}{\kappa_k^*}
         \left[\kappa \bm + \sum_{i=1}^N r_{i,k}\bx_i   \right]
         \\
         \nu_k^* &= \nu + \sum_{i=1}^N r_{i,k} 
         \\
         \bPsi_k^*&=
         \frac{1}{\nu^*}
         \left[
         \nu \bPsi +\kappa \bm\bm^T 
         +\left(\sum_{i=1}^N r_{i,k}\bx_i\bx_i^T\right)- \kappa_k^*\bm_k^*(\bm_k^*)^T 
         \right] \ .
        \end{align}

The parameters of the subclusters are updated in a very similar way, except that the soft subcluster assignments (\ie, $(\widetilde{r}_{i,j})$) are used instead of the soft cluster assignments.

\subsection{The Concentration Parameter of the Dirichlet Process}\label{Sec:DP}
The concentration parameter of the Dirichlet Process, $\alpha>0$, is a user-defined hyperparameter
that, when sampling from the \emph{prior}, controls (the expected number of) the number of clusters. 
In short, the higher $\alpha$ is, the more clusters are expected. 
However, when doing \emph{posterior calculations}, if $\alpha \ll N$, where $N$ is the number of data points, then the importance of $\alpha$ diminishes. 
Particularly, when computing the Hastings ratios for the splits and merges, the importance of $\alpha\ll N$ is usually negligible.

\section{The Factors Affecting $K$; Our Weak Prior}
In a DPGMM, the number of clusters is affected not only by $\alpha$ 
and the data but also the NIW prior. For example, if $\nu$ is very high
and $\bPsi$ is small, then the model will favor small clusters
and thus $K$ is likely to be high (so the small clusters will efficiently cover the data). Conversely, if $\nu$ is very hight and $\bPsi$ is large, 
then the model will favor large clusters so $K$ will tend to be small (only few large clusters can cover the entire data). 
However, we emphasize that in all cases, our choices (see below) for the values of the NIW hyperparameters and the concentration parameter $\alpha$  
correspond to a very weak prior. That is, our $\alpha$, $\nu$ and $\kappa$ are all much smaller than $N$ in all the datasets we experimented with:
across all the datasets, the smallest $N$ was 13,000 (in STL-10).
As a result, the main factor in determining the inferred $K$ is the data itself. 

\section{Merge Proposals}
In the paper, we explained how $K$ is changed via splits and merges, and described how to compute the acceptance probability of split proposals. Here, we provide the complementary details on merge proposals. In a merge step, we sequentially propose pairs of clusters to merge. To avoid sequentially considering all possible merges, we consider
(sequentially) the merges of each cluster with only its 3 nearest neighbors.

The proposal to merge a pair of clusters, $k_1$ and $k_2$, is a accepted with probability (1, $H_\mathrm{m}$), where
\begin{align}
	H_{\mathrm{m}} = \frac{1}{H_{\mathrm{s}}} = \frac{\Gamma(N_{k_1} + N_{k_2}) f_\bx(\Xcal_{\{k_1,k_2\}};\lambda)}{\alpha \Gamma(N_{k_1}) f_\bx(\Xcal_{k_1};\lambda)  \Gamma(N_{k_2}) f_\bx(\Xcal_{k_2};\lambda)}
\end{align}

is the Hastings ratio, $\Gamma$ is the Gamma function, $N_{k,1}$
and $N_{k,2}$ are the number of points in clusters $1$ or $2$, respectively, 
$\Xcal_k=(\bx_i)_{i:z_i=k}$ denotes the points in cluster $k$, and  $\Xcal_{\{k_1, k_2\}}=(\bx_i)_{i:z_i \in \{k_1, k_2\}}$ denotes the points in clusters $k_1$ and $k_2$. As for $f_\bx(\cdot;\lambda)$, this is the \emph{marginal} likelihood where $\lambda$
represents the NIW hyperparameters.

\section{Feature Extraction}
Here we provide more details on the feature-extraction process. In general, there are two main paradigms: an end-to-end approach in which the features and clustering are learned simultaneously, and a two-step approach in which clustering is performed on pre-computed latent features.\\

\subsection{End-to-end Approach: Jointly Learning Features and Clustering} Here, we loosely follow DCN~\cite{Yang:ICML:2017:towards} and similarly start by training an Autoencoder (AE) with a reconstruction loss, 
\begin{align}
	\Lcal_{\mathrm{AE_{recon}}} = \tfrac{1}{N}\sum\nolimits_{i=1}^N \ellTwoNorm{\bg(\boldf(\bx_i)) - \bx_i}^2
 	\end{align}
where $\boldf$ is the encoder and $\bg$ is the decoder. Then, while DCN performs $K$-means on the resulted embeddings to obtain initial cluster centers and assignments, we use our more flexible DeepDPM (which, unlike $K$-means, assumes neither isotropic classes, uniform weights, nor a known $K$). Next, we utilize the pipeline of~\cite{Yang:ICML:2017:towards} and refine the AE's latent space by training it with both $\Lcal_{\mathrm{recon}}$ and an additional clustering loss:
  \begin{align}
 	\Lcal_{\mathrm{AE_{clus}}} = \ellTwoNorm{\boldf(\bx_i) - \bmu_{z_i}}^2,
  \end{align}
 where $z_i$ is the cluster assignment of $\bx_i$, and $\bmu_{z_i}$ is the cluster's center. This loss encourages $\boldf$ to create small intra-class distances in the latent space. The overall loss is $ \Lcal_{\mathrm{AE_{recon}}} + \tfrac{\beta}{2}\Lcal_{\mathrm{AE_{clus}}}$
where $\beta>0$ is a tunable parameter. While in~\cite{Yang:ICML:2017:towards} new cluster assignments and centers are computed after each epoch of the AE, we keep them fixed during this stage, changing only the embeddings. While the pipeline suggested in~\cite{Yang:ICML:2017:towards} ends here, we add an alternation scheme where we alternate between clustering the updated embeddings using DeepBNP (keeping the AE frozen) and training the AE (\ie, perform feature learning, while keeping the clustering fixed). We repeat this process several times.
Intuitively,
during training, DeepDPM
is likely to change $K$, thus, adapting the embeddings accordingly may reveal inter- and intra-class structures which can be useful, in turn, for the clustering module to find meaningful clusters.\\ 

\subsection{Two-Step Approach: Training on Latent Features.} Another approach for deep clustering is using a pretrained feature extractor backbone. As our method is DL-based, it is easy to concatenate any DL backbone before our DNN. Thus, we follow~\cite{Gansbeke:ECCV:2020:SCAN} and use MoCo~\cite{DC:Chen:2020:MoCO} for (unsupervised) feature extraction.

We provide the specific values we used for the feature extractions below in~\autoref{Sec:Supmat_implementation}.

\section{Implementation Details and Hyperparameters}\label{Sec:Supmat_implementation}
We detail here all the training configurations and hyperparameters used in our experiments.

For all the experiments, our clustering net used the following MLP architecture where the MLP had an input layer, a single hidden layer, and an output layer. The number of neurons in the input layer was $d$ (the dimension of the input to the clustering module). The number of hidden units was always 50 in all our experiments (changing that number had little effect on the results). The (changing) number of neurons in the output layer was $K$ (which corresponds to the changing number of clusters).  In addition, in all our experiments we used a batch size of 128, a learning rate (lr) of 0.0005 for the clustering net, and an lr of 0.005 for the subclustering nets. As for the prior hyperparams, for the DP's $\alpha$ we chose $\alpha = 10$, and for the NIW hyperparams we used $\kappa = 0.0001$, set $\bm$ to be the data mean, and $\nu$ to be $d + 2$. We used different $\bPsi$ values in each experiment, as detailed below.

 Below in~\autoref{intuition} we give some guidelines on how to choose the key hyperparameters for DeepDPM.

\subsection{Setup Used for Comparing with Classical Methods}

\textbf{Feature Extraction.} For this experiment, we used the feature extraction procedure suggested by~\cite{DC:Mcconville:ICPR:2021:n2d} where we first trained a deep Autoencoder (AE) and then transformed its latent space using a UMAP transformation~\cite{DC:Mcinnes:2018:umap}. We used the same configurations as in~\cite{DC:Mcconville:ICPR:2021:n2d}.

\textbf{DeepDPM hyperparameters.} For all experiments, we initialized DeepDPM with $K=1$, DeepDPM was trained for 500 epochs. We set $\bPsi=\bI \times 0.005$ in all the datasets, where $\bI$ denotes the identity matrix.
Note that we used the same configuration for the three datasets, in both the balanced and imbalanced cases.

\subsection{Training on Latent Features}
Here, we give the hyperparameters we used for evaluating our method on STL-10 and ImageNet. 
For both datasets we used the unsupervised pretrained feature extractor MoCo~\cite{DC:Chen:2020:MoCO} and trained DeepDPM on top of the resulting features. For STL-10 we pretrained it for 1000 epochs (on STL-10's train set) and for ImageNet we used the pretrained weights available online.\\

For STL-10, we initialized DeepDPM with $K = 3$ and trained it for 500 epochs with $\bPsi= \bI \times 0.05$. For ImageNet, we initialized DeepDPM with $K = 150$ and trained it for 200 epochs with $\bPsi= \bI \times 0.001$.

\subsection{Jointly Learning Features and Clustering}
\begin{table*}
\centering
\begin{tabular}{l ccc}
\hline
\textbf{Dataset} & AE architecture & AE lr & DeepDPM training epochs\\
\hline 
\hline
MNIST & D-500-500-2000-10 & 0.002& 200\\
Reuters10K & D-500-500-2000-75&0.002&300\\
ImageNet-50& D-500-500-2000-10 & 0.002 & 300\\
\hline
\end{tabular}
\caption[]{Implementation details for DeepDPM's experiments. $D$ denotes the input dimension. For all datasets but ImageNet-50 it is the original data dimension. For ImageNet-50, $D = 128$, the output dimension of MoCo~\cite{DC:Chen:2020:MoCO}}
\label{results:implementation_dets}
\end{table*}

As described in the paper, we adapted the feature learning pipeline from DCN~\cite{Yang:ICML:2017:towards} to jointly learn features and clustering in alternation. 
\autoref{results:implementation_dets} shows the AE architectures,  the lr values, and the number of DeepDPM epochs. For ImageNet-50, we trained an AE on top of the features generated by MoCo~\cite{DC:Chen:2020:MoCO}. For all other datasets, it was trained on top of the original data dimension. For MNIST and Reuters10K we used three alternations, for ImageNet-50 we used two alternations in the balanced case and four alternations for the imbalanced. See below in~\autoref{intuition} how we chose the number of alternations.\\

DeepDPM was initialized with $K = 3$ for MNIST, $K = 1$ for Reuters10K and $K = 10$ for ImageNet-50. As per the prior hyperparams, we chose $\bPsi = 0.005$ for MNIST and Reuters10K. For ImageNet-50 we chose $\bPsi= \bI \times \mathrm{std}(\Xcal) \times 0.0001$ , where $std(\Xcal)$ denotes the standard deviation of the data.
Note that in both the balanced and imbalanced cases of ImageNet-50 we used the same hyperparameters, where the only difference was the number of alternations.

\subsection{General Recommendations for Choosing Hyperparameters}\label{intuition}
\textbf{Number of epochs.} The number of epochs is chosen by the amount of epochs after which DeepDPM has converged to a certain $K$; \ie, no more split/merge proposal are accepted. We chose it empirically by training DeepDPM and measuring the average number of epochs it took for $K$
to stabilize. We set the maximal number of epochs to 100 plus that average epoch number. 

\textbf{Number of alternations.} When performing feature learning and clustering in alternation, we need to choose the number of times we perform the alternations (one alternation includes training the AE followed by the DeepDPM training). We stop the alternations
once the DeepDPM's inferred $K$ stabilizes.  

\textbf{The initial value for $K$.} As we showed, the initial value of $K$ has little effect on the final clustering that DeepDPM generates. Thus, it can be chosen arbitrarily. That said, the value of the initial $K$ can affect the speed of convergence (if DeepDPM starts with a more accurate estimate
then less splits and merges will happen and the DeepDPM will stabilize faster). A reasonable choice is to choose the initial $K$ to be proportional to $N$, the number of datapoints; \eg, $N / 10000$. Note that unlike parametric methods, \emph{this is used only as an initialization value which is expected to change}. 

\textbf{Choosing $\bPsi$.} As a rule of thumb, we took $\bPsi$ to be proportional to $\bI$ (\ie, the $d\times d$ identity matrix) with the elements of the main diagonal being the data's standard deviation, scaled by an additional scalar $s$. The choice of $s$ is based on empirically seeing that a substantial amount of both splits and merges proposals are accepted during the first few hundred epochs. Note that as $\bPsi$ is getting smaller, more splits will be accepted. Thus, if $\bPsi$ is too small, only splits will occur (and no merges) and if it is too large, no splits will be accepted.

\clearpage
{\small
\bibliographystyle{./ieee_fullname}
\bibliography{./refs}
}